\documentclass[10pt, twocolumn, letterpaper]{article}

\usepackage[pagenumbers]{goodtwocolumn} %

\usepackage[utf8]{inputenc}		%
\usepackage[T1]{fontenc}    	%
\usepackage{amsfonts}       	%
\usepackage{nicefrac}       	%
\usepackage{microtype}      	%
\usepackage[dvipsnames]{xcolor} %

\usepackage{graphicx}
\usepackage{amsmath}
\usepackage{amssymb}

\usepackage{subcaption}     %
\usepackage{multirow}
\usepackage{tabularx}
\usepackage{booktabs}

\usepackage{footnote}
\usepackage{chngpage}
\usepackage{placeins}
\usepackage{ragged2e}
\usepackage{soul}
\usepackage{multirow}
\usepackage{graphicx}

\usepackage[sort, numbers]{natbib}

\usepackage[pagebackref, breaklinks, colorlinks]{hyperref}
\usepackage{url}            %

\usepackage[capitalize]{cleveref}
\crefname{section}{Sec.}{Secs.}
\Crefname{section}{Section}{Sections}
\Crefname{table}{Table}{Tables}
\crefname{table}{Tab.}{Tabs.}

\hypersetup{
  colorlinks,
  citecolor=Green,
  linkcolor=RoyalBlue,
  urlcolor=RoyalBlue}
\begin{document}

\title{Let's Go Shopping (LGS) --\\Web-Scale Image-Text Dataset\\for Visual Concept Understanding}

\author{
\normalsize{Yatong Bai$^{1*}$ \quad Utsav Garg$^2$ \quad Apaar Shanker$^2$ \quad Haoming Zhang$^2$ \quad Samyak Parajuli$^2$}\\
\normalsize{Erhan Bas$^2$ \quad Isidora Filipovic$^3$ \quad Amelia N. Chu$^3$ \quad Eugenia D Fomitcheva$^3$ \quad Elliot Branson$^2$}\\
\normalsize{Aerin Kim$^2$ \quad Somayeh Sojoudi$^1$ \quad Kyunghyun Cho$^3$}\\[3mm]
$^1$University of California, Berkeley $\qquad^2$Scale AI $\qquad^3$New York University\\
{\small $^*$Work done during internship at Scale. \quad
Correspondences to \texttt{\href{mailto:<yatong_bai@berkeley.edu>?Subject=Your CVPR 2023 paper}{yatong\_bai@berkeley.edu}{}, \href{mailto:aerinykim@gmail.com>?Subject=Your CVPR 2023 paper}{aerinykim@gmail.com}{}}.}
}

\maketitle

\begin{abstract}
Vision and vision-language applications of neural networks, such as image classification and captioning, rely on large-scale annotated datasets that require non-trivial data-collecting processes. This time-consuming endeavor hinders the emergence of large-scale datasets, limiting researchers and practitioners to a small number of choices.
Therefore, we seek more efficient ways to collect and annotate images. Previous initiatives have gathered captions from HTML alt-texts and crawled social media postings, but these data sources suffer from noise, sparsity, or subjectivity. For this reason, we turn to commercial shopping websites whose data meet three criteria: cleanliness, informativeness, and fluency.
We introduce the Let's Go Shopping (LGS) dataset, a large-scale public dataset with 15 million image-caption pairs from publicly available e-commerce websites. When compared with existing general-domain datasets, the LGS images focus on the foreground object and have less complex backgrounds.
Our experiments on LGS show that the classifiers trained on existing benchmark datasets do not readily generalize to e-commerce data, while specific self-supervised visual feature extractors can better generalize. Furthermore, LGS's high-quality e-commerce-focused images and bimodal nature make it advantageous for vision-language bi-modal tasks: LGS enables image-captioning models to generate richer captions and helps text-to-image generation models achieve e-commerce style transfer.
\end{abstract}

\section{Introduction}

\begin{figure*}[!tb]
    \centering
    \begin{minipage}{.375\textwidth}
        \frame{\includegraphics[width=\textwidth, trim={15mm, 3mm, 15mm, 5mm}, clip]{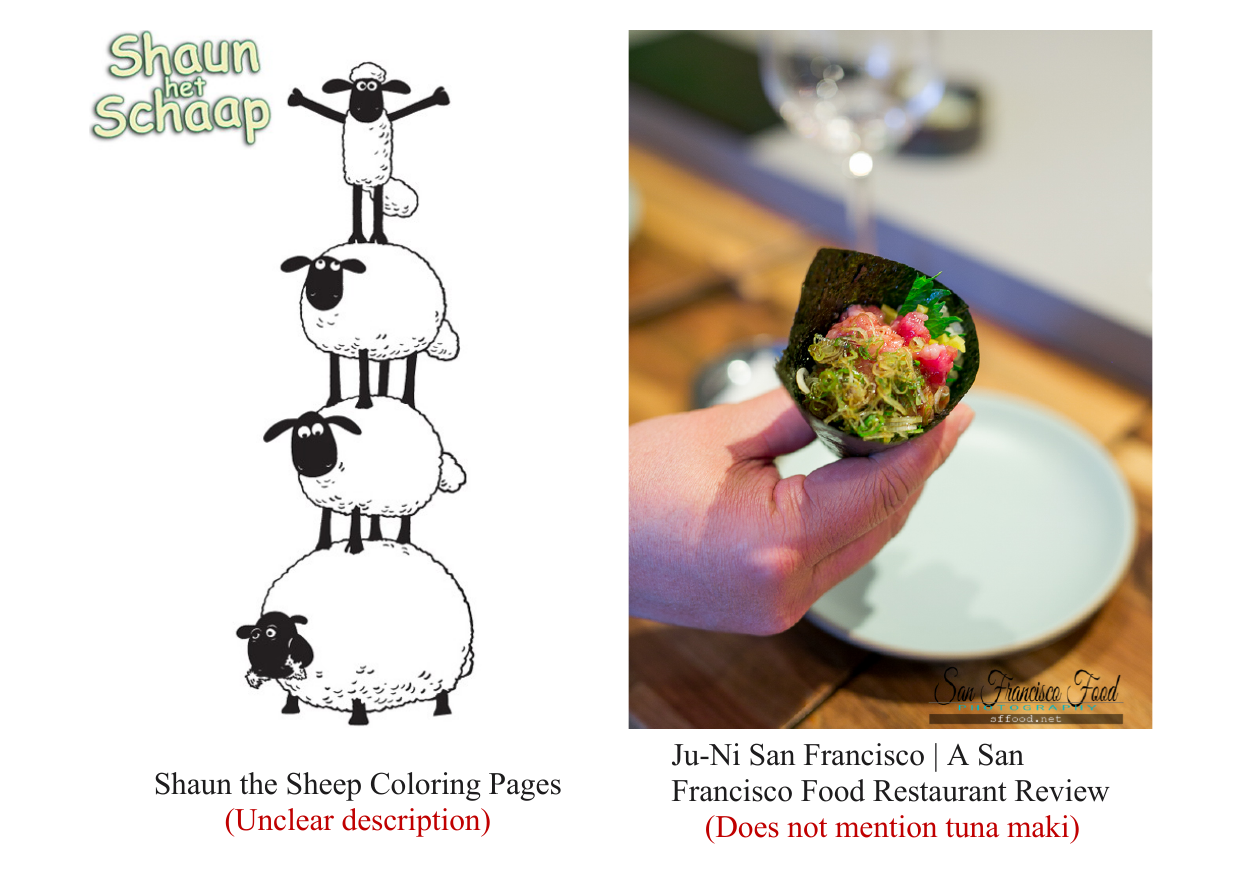}}
        \caption{In comparison to e-commerce product descriptions, alt-text is usually less informative, sometimes too broad, or even irrelevant.}
        \label{fig:sample_description_alt}
    \end{minipage}
    \hfill
    \begin{minipage}{.615\textwidth}
    	\includegraphics[width=\textwidth, trim={44mm, 3cm, 46mm, 3cm}, clip]{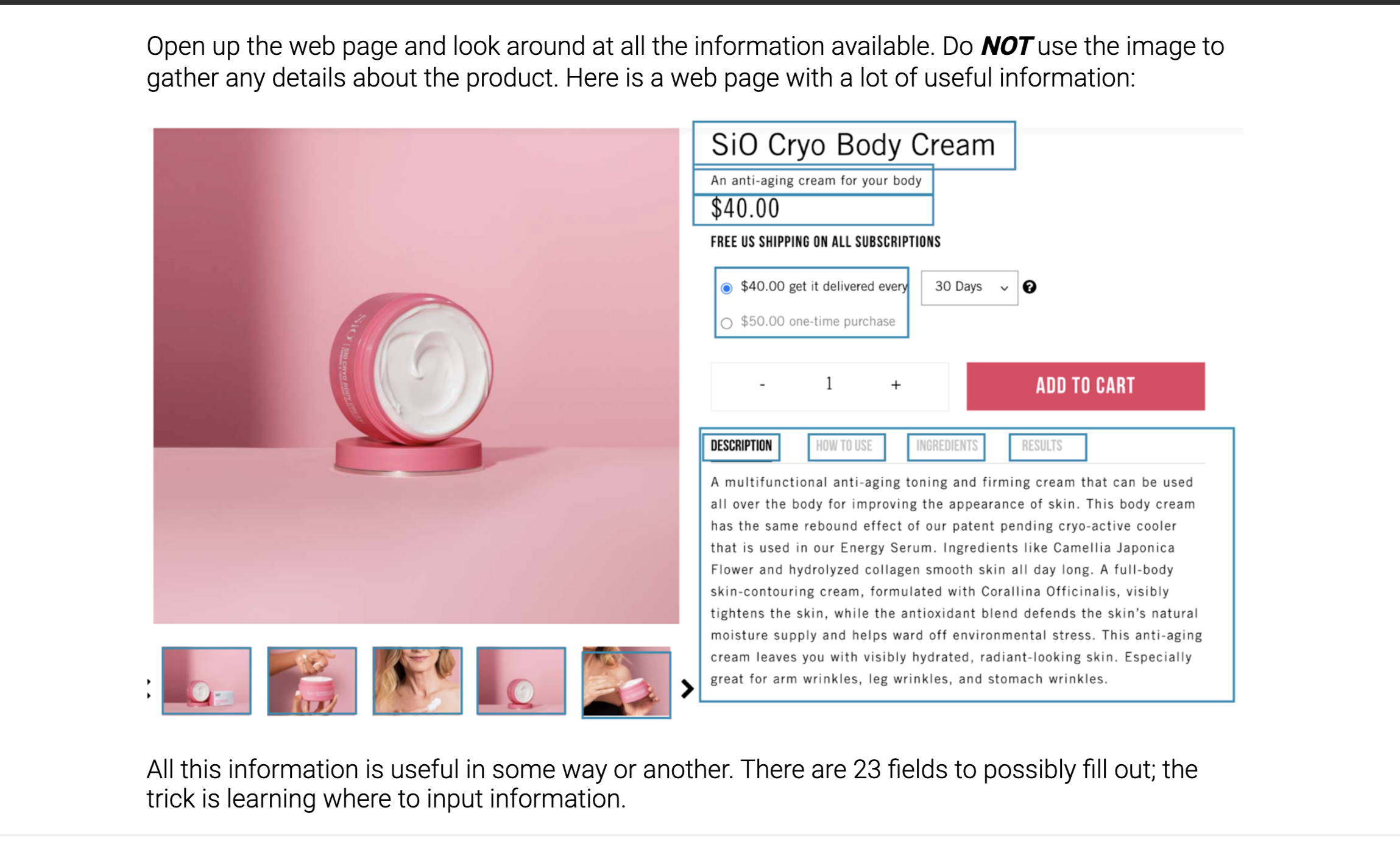}
    	\vspace{-5.5mm}
    	\caption{An e-commerce-based LGS sample instance with image, title, and description.}
    	\label{fig:sample_description_lgs}
    	\vspace{3mm}
	\end{minipage}
\end{figure*}

Computer vision (CV) and natural language processing (NLP) tasks increasingly rely on pre-trained representations. While NLP representations can be trained on unannotated raw text, vision applications often consider pre-training using large-scale datasets with discrete class labels annotated by humans, such as ImageNet \cite{russakovsky2015imagenet, deng2009imagenet} or OpenImages \cite{kuznetsova2020open}. Vision-language bimodal applications, such as image captioning or visual question answering, similarly rely on large amounts of annotated data. Unfortunately, many of the large-scale bi-modal datasets now in existence, such as CLIP \cite{radford2021clip}, ALIGN \cite{jia2021scaling}, and JFT300M \cite{hinton2015distilling, chollet2017xception}, are not publicly accessible. As a result, research has been constrained to a few selected large datasets, such as Conceptual Captions \cite{changpinyo2021conceptual} and COCO \cite{chen2015microsoft}. This shortage of available public datasets can be attributed in part to the time and effort required to gather, clean, and annotate large datasets.

Therefore, we adopt a more efficient and scalable high-quality data collection pipeline to acquire image-text pairs easily available on e-commerce websites. While some existing datasets use public websites as annotation sources, most of them use social media websites (RedCaps \cite{Desai2021RedCapsWI}) or alt-texts\footnote{Alt-texts are short descriptions of HTML website images. When an image cannot be rendered, the website displays its alt-text as a surrogate.} (Conceptual Captions \cite{sharma2018conceptual}) for this purpose. Nevertheless, social media data suffer from subjectivity. On the other hand, alt-texts can be unacceptably noisy, sometimes merely including uninformative texts such as ``alt img'', as shown in \Cref{fig:sample_description_alt}. 

As a result, we gravitate to e-commerce websites, where clean images with objective, accurate, succinct, and informative descriptions are abundant, as illustrated in  \Cref{fig:sample_description_lgs}. Let's Go Shopping (LGS) dataset collects 15 million image-description pairs from approximately 10,000 e-commerce sites selling a wide range of products. Due to the nature of e-commerce data, the majority of LGS images have a clear background and a static focus on the stated object. On the captions front, LGS provides precise and elaborative captions. We show how highly precise information can be extracted from captions for vision-language fine-tuning. 

On the other hand, ImageNet-1k has served as the ubiquitous go-to pre-training and evaluation dataset for vision-only applications. While ImageNet covers a wide range of domains, the diversity of angles and arrangements is restricted. As a result, the literature has shown that ImageNet models do not generalize well to deliberately constructed out-of-distribution (OOD) scenarios \cite{Andrei2019ObjectNet}. This work uses image classification experiments to demonstrate that such OOD data is ubiquitous in e-commerce applications. We then show that models can benefit from the unique e-commerce distribution in image classification, reconstruction, captioning, and generation tasks. 

Specifically, we convert the LGS captions into taxonomies and labels and demonstrate a large disparity between the label distributions of LGS and ImageNet: even with best efforts, only 17.6\% of the concepts are shared between popular ImageNet-1k synsets and the e-commerce corpus (more details in \Cref{sec:exp_classi}). Even for those shared classes, the performance of ImageNet models degrades significantly. By verifying that the LGS classes are well-separable, we conclude that this performance degradation can be mostly attributed to the distributional disparity. To separate the effects of labels and captions and isolate the distribution shift of the images, we consider Masked AutoEncoder (MAE) \cite{He2022MAE}, a self-supervised pre-training method that does not rely on labels. We show that an MAE model trained on ImageNet-1k can reconstruct LGS images well, but adding LGS to the training data improves the performance on LGS and generalizes better to COCO.

The above results demonstrate that while the e-commerce images are from a distribution that is distinct from current benchmark datasets, the feature extractors can be shared.
Moreover, we illustrate additional merits of LGS that qualify it as a pre-training dataset. Specifically, the models learned on both LGS and ImageNet have improved linear probing performance on common downstream tasks such as CIFAR-100 \cite{cifar10} and Fashion MNIST \cite{Xiao17FMNIST}, compared with the ImageNet-only counterparts. 

The distinctive distribution of LGS also benefits vision-language bimodal tasks. For caption generation tasks, we train an OFA model \cite{wang2022OFA} on LGS to demonstrate that the more prominent image foreground, cleaner image background, and the highly descriptive captions of LGS enable the model to produce ``attribute-rich'' image captions, which models trained on traditional datasets fail to produce.

Finally, for text-to-image generation tasks, diffusion models \cite{diffusion, ddpm, constta} form the currently most popular family of methods. To illustrate the efficacy of LGS in this setting, we use Stable Diffusion (SD) \cite{rombach2021highresolution} and fine-tune it in both general and fine-grained settings on subsets of the LGS dataset. We demonstrate promising qualitative and quantitative results on adapting existing text-to-image models using LGS for e-commerce-related generations. Furthermore, with the help of its distinct image style and descriptive captions, LGS can help the SD model generate e-commerce-styled images.

To make LGS available to the public, we will share the filtered links to the image-caption pairs under the ``BSD 3-Clause'' license (also used in common datasets such as ImageNet), as was the case for ImageNet. We will also share the downloader so that the exact same dataset can be reproduced.

\section{Related Work} \label{sec:related_work}

\subsection{Unimodal Pre-Training Datasets}

Prior to the popularization of bi-modal training, uni-modal data (vision-only or language-only) have been the workhorses for pre-training tasks. On the vision side, ImageNet-1k and ImageNet-22k are still some of the most prevalent examples, alongside the larger JFT-300M dataset. For the e-commerce domain, Fashion MNIST, Clothing1M \cite{xiao2015learning}, Fashion200k \cite{han2017automatic}, and FashionIQ \cite{guo2019fashion} have been proposed to analyze the effects of noisy labels. Some of the most common datasets used as general wide-domain downstream tasks include CIFAR-10, CIFAR-100, MNIST \cite{lecun2010mnist}, SVHN \cite{SVHN}, and Tiny ImageNet \cite{Le2015TinyIV}.

\subsection{Vision-and-Language Pre-Training Datasets}

The literature has shown that image-text data from COCO can be used to learn \emph{visual} features that are competitive with supervised pre-training \cite{he2016deep} on ImageNet when transferred to downstream tasks \cite{everingham2009voc, zhou2014places, lin2014microsoft, gupta2019lvis, van2018inaturalist, desai2020virtex, bulent2020icmlm}. More recently, CLIP and ALIGN scaled up to 400M and 1B+ web-curated image-text pairs, enabling zero-shot visual recognition on downstream tasks.

\begin{table}[!tb]
    \centering
    \caption{The instance count of LGS compared with existing bi-modal datasets.}
    \begin{small}
    \scalebox{.91}[1.]{\begin{tabular}{lc} 
        \toprule
        \textbf{Datasets} & \textbf{Instances} \\
        \midrule
        Let's Go Shopping \scriptsize{(this paper)} & 14,847,764 \\
        YFCC100M \scriptsize{(Yahoo)} & 100 million \\
        RedCaps \scriptsize{(University of Michigan)} & 12,011,111 \\
        Conceptual Captions 12M \scriptsize{(Google)} & 12,423,374 \\
        WIT-English \scriptsize{(Google)} & 5,500,746 \\
        Localized Narratives \scriptsize{(Google)} & 849,000 \\
        COCO \scriptsize{(Microsoft)} & 328,000 \\
        Visual Genome \scriptsize{(Stanford)} & 108,077 \\
        CLIP \scriptsize{(OpenAI)} & 400M \\
        ALIGN \scriptsize{(Google)} & 1.8B \\
        \bottomrule
    \end{tabular}}
    \end{small}
    \label{tab:dataset_size}
\end{table}

Originally intended for image-text retrieval and image captioning,
bi-modal datasets are now widely used for training cross-modal representations \cite{tan2019lxmert, lu2019vilbert, li2019visualbert, su2019vl, li2020unicoder, chen2019uniter, zhou2020vlp, li2020oscar, huang2020pixelbert, kim2021vilt, ordonez2011im2text, sharma2018conceptual}
that transfer to downstream tasks, such as visual question answering \cite{antol2015vqa, zhu2016visual7w, hudson2019gqa}, referring expressions \cite{kazemzadeh2014referitgame}, and visual reasoning \cite{suhr2019corpus, zellers2019recognition}.
In light of these novel training paradigms, more recent works build larger datasets specifically for vision-and-language pre-training. Examples include LAIT \cite{qi2020imagebert}, Conceptual Captions-12M, and Wikipedia-ImageText (WIT) \cite{srinivasan2021wit}, Localized Narratives \cite{PontTuset_eccv2020}, Visual Genome \cite{krishna2017visual}, YFCC100M \cite{yfcc100m}.
Similar to these datasets, LGS offers rich semantic data for pre-training applications. However, our choice of e-commerce data source is unique, leading toward distinctive data distribution.

Image-text datasets are also used for learning visual features.
The work \cite{li2017ngrams} has proposed to train visual $n$-gram models on YFCC100M, whereas other methods
\cite{desai2020virtex, bulent2020icmlm} aim to learn features from the captions from the COCO dataset~\cite{chen2015microsoft}. The quality of the resulting features is competitive with supervised ImageNet training~\cite{he2016deep} on many downstream tasks~\cite{everingham2009voc, russakovsky2015imagenet, lin2014microsoft, gupta2019lvis, van2018inaturalist}. Moreover, the image-text pre-training schemes scale up to very larger non-public datasets that are even larger than LGS \cite{radford2021clip, jia2021scaling}.

A core motivation for collecting image-text pairs from the internet is the possibility of scaling up the data size without bearing the prohibitively expensive annotation costs.
In light of this motivation, there have been multiple efforts to collect large quantities of noisy labels associated with online images, leading to datasets such as WebVision \cite{li2017webvision}, YFCC100M, JFT-300M, and Instagram-3.5B \cite{mahajan2018exploring}.

Existing multi-modal e-commerce-inspired datasets include M5Product \cite{dong2022m5product} and DeepFashion \cite{liu2016deepfashion}. With 6 million instances, M5Product's size is around half of LGS's. While M5Product focuses on demonstrating the effectiveness of multi-modal training, this paper emphasizes analyzing the e-commerce data distribution and how it generalizes to general wide-domain datasets in a pre-training setting.

\section{The Let's Go Shopping (LGS) Dataset} \label{sec:dataset_analysis}

With 14,847,764 image-text pairs, the LGS dataset has a size advantage over many publicly available bi-model datasets, as presented in \Cref{tab:dataset_size}. In this section, we offer additional analysis of the LGS data. For all analysis and experiments in the paper, we use a subset of the instances with 13 million instances, as the rest of the dataset was constructed in parallel with the experiments.

\subsection{Data Collection} \label{subsec:data_collection}

To create training data that is truly representative of e-commerce data as a whole, we include a wide range of commerce websites with various product kinds, such as infant products, sporting goods, bridal jewelry, etc.

The collection pipeline starts with a set of heuristic rules to isolate the product pages from the non-product pages of an e-commerce website. Then, our automated extractor obtains relevant information on each product page, including the product title, the description, and the first listed image. Some products may include numerous variants (e. g., different colors for a type of T-shirt), and we collect all variants. We avoid crawling information that the sellers are unwilling to share. Specifically, the extractor is forbidden from crawling pages with a ‘Disallow’ extension.
Finally, we use strict automated tests to filter out the instances with potential quality issues. Examples of the tests include confirming that the price is a number, certifying that the images are valid, and ensuring that the product title exists and contains no unexpected characters.

\subsection{Characteristics of LGS Images}

\begin{figure*}
    \centering
    \includegraphics[width=\textwidth]{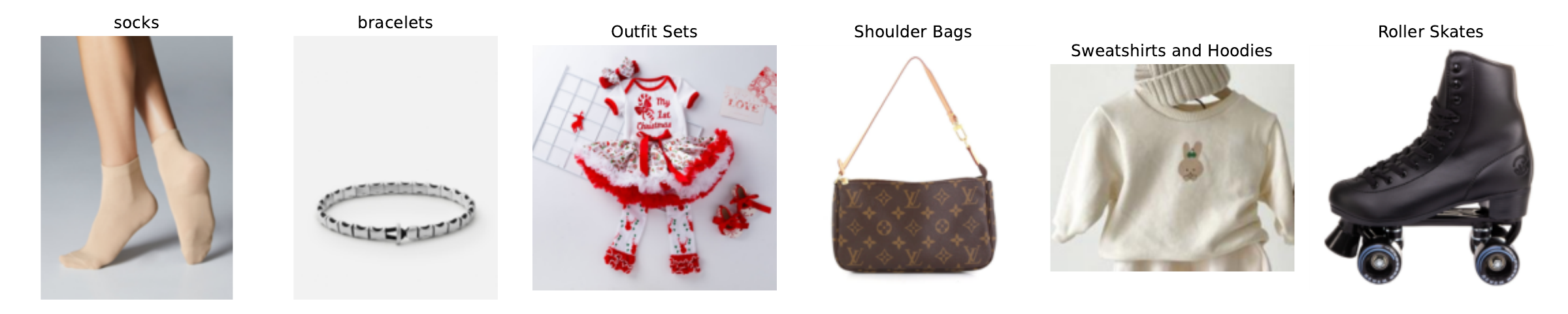}
    \vspace{-7mm}
    \caption{Examples of LGS images with taxonomy end leaves}
    \label{fig:lgs_img_exp}
\end{figure*}

In general-domain image-caption datasets, the images usually consist of one or more subjects juxtaposed against a rich background, and their captions often mention the background. In contrast, e-commerce product thumbnails in LGS often depict only one in-animate item that occupies the foreground without any association with the background. The background is also often a single color, with some examples shown in \cref{fig:lgs_img_exp}. These clear backgrounds make it easier for models to locate the patterns that correspond to their tasks.

\subsection{Characteristics of LGS Captions} \label{sec:lgs_captions}

In this subsection, we analyze the traits of the LGS captions. The LGS dataset has 14,847,764 captions in total, and the words and phrases in LGS captions are diverse. For example, while LGS has around 3x more captions than COCO\footnote{Each COCO instance has five corresponding captions, and we consider each of them separately.}, its captions possess about 20x more uni-grams, bi-grams, and tri-grams, with more detailed statistics presented in \Cref{sec:ngrams}. \Cref{tab:word_stats} presents some statistics of the word distribution of the captions, showing that both LGS and COCO have highly positively skewed distributions, with LGS having a longer tail. Since LGS incorporates data from a large variety of e-commerce websites, the descriptions can include rich information. In the subsequent sections, we show that while the raw captions of LGS are diverse, clear structural information can be extracted from the LGS captions for fine-tuning purposes.

\begin{table}[!tb]
  \begin{center}
    \caption{Comparing the word count statistics of the LGS and COCO captions.}
    \begin{small}
    \begin{tabular}{l !{\vrule width 2pt} c|c|c|c|c}
      \toprule
      \textbf{Dataset} & \textbf{Min} & \textbf{Max} & \textbf{Mean} & \textbf{Median}  & \textbf{Skew} \\
      \midrule
      LGS  & 2 & 3642 & 89.58 & 67 & 3.44 \\
      COCO & 5 & 50   & 10.56 & 10 & 2.76 \\
      \bottomrule
    \end{tabular}
    \end{small}
    \label{tab:word_stats}
  \end{center}
\end{table}

\begin{table}[!tb]
  \begin{center}
    \begin{small}
    \caption{The POS's that occur at least ten times.}
    \scalebox{.92}[1.]{\begin{tabular}{l !{\vrule width 2pt} c|c|c|c}
      \toprule
      \textbf{Dataset} & \textbf{C. Nouns} & \textbf{P. Nouns} & \textbf{Adjectives} & \textbf{Verbs} \\
      \midrule
      LGS  & 158,479 & 139,174 & 48,907 & 57,481 \\
      COCO & 10,403  & 1,655   & 3,053  & 4,961  \\
      \bottomrule
    \end{tabular}}
    \end{small}
    \label{tab:pos_stats}
  \end{center}
\end{table}

Additionally, we use the part-of-speech (POS) tagging method from the Spacy library \cite{Honnibal20} to analyze the linguistic statistics of the LGS captions, comparing common nouns, proper nouns, adjectives, and verbs. \Cref{tab:pos_stats} illustrates that LGS has at least 10x more words per POS compared with COCO, whereas Figures \ref{fig:lgs_pos} and \ref{fig:coco_pos} in the supplementary materials provide further insights into the composition of each word type. Due to the e-commerce nature of LGS, a large portion of the instances is clothing and other wearable items. Thus, within LGS, the proper nouns often present the brand names and sizes, the common nouns often describe the materials, and the adjectives and verbs often characterize the product-specific descriptions and actions, making the LGS captions highly descriptive.

\subsection{LGS for Classification} \label{sec:exp_classi}

While the raw data format of LGS is image-caption pairs, we also experimented with image classification with LGS by labeling the classes. Specifically, we build three classification variants: LGS-117, LGS-710, and LGS-Overlap. For all three variants, we use a taxonomy generation language model pre-trained in-house to convert each product description into a taxonomy tree, whose nodes are designed to be informative for e-commerce catalog applications. The end leaf of each taxonomy tree is then used as the label, with some examples displayed in \Cref{fig:lgs_img_exp}. The taxonomy tree can also be used to generate summarized image captions that include \textit{product title}, \textit{product brand name}, and a number of \textit{``bullet strings''} describing specific product attributes. The bullet strings include examples such as \texttt{Nylon fabric}, \texttt{Classic collar}, and \texttt{Front zipper fastening}. The LGS leaves form a long-tailed distribution that emphasizes common daily commodities, with the five most common leaves being \texttt{Tops and T-shirts}, \texttt{Dresses}, \texttt{Rings}, \texttt{T-shirts}, and \texttt{Sweatshirts and Hoodies}. For each of the three classification variants, we further clean the end leaves, with details provided in the two following paragraphs. In \Cref{fig:lgs_class_dist} in the supplementary materials, we provide a histogram of the end leaf distribution.

LGS-117 and LGS-710 are designed as pre-training datasets. Within all raw labels generated by the taxonomy model, there are synonyms and overlaps that should be unified. After manually merging the synonyms among the most popular classes, we observe 117 classes that contain at least 10k images. We select 10k images from each class, forming the balanced LGS-117 dataset. LGS-710 is an unbalanced dataset that includes more scarce classes. To accelerate label engineering, we use a semi-automated pipeline. First, we remove uninformative words like ``other'' and parse juxtaposed nouns by commas and ``and''. Next, we use a pre-trained language model to extract the embedding of each parsed noun. As an example, for the leaf \texttt{Tops and T-shirts}, we embed both \texttt{tops} and \texttt{t-shirts}. We then consider the ``similarity'' between two classes to be the maximum cosine similarity between all pairs of corresponding nouns. Very close classes are merged based on a similarity threshold of 0.92, which is determined by manually inspecting the merged classes.

LGS-Overlap is proposed as an out-of-distribution test set for models trained on ImageNet-1k, one of the most widely used benchmarking datasets. We use a similar semi-automated pipeline to merge LGS classes with ImageNet synsets \cite{WordNet, deng2009imagenet}. We optimize the pipeline by adjusting the similarity threshold to 0.90 and including additional pre-processing steps such as singularization and keyword merging. Note that polysemous words in the labels can refer to different objects in LGS and ImageNet. For example, ``cricket'' in LGS refers to sports equipment but refers to the insect species in ImageNet. Thus, a manual inspection of the merged classes is performed. After discarding classes with less than 20 instances, we gather the remaining 176 ImageNet synsets that align with the LGS end leaves and use them as the LGS-Overlap dataset. The fact that only 17.6\% of the ImageNet synsets are matched shows a significant label distribution difference between e-commerce applications and common pre-training datasets. Since a higher level of label-space alignment is essential for more effective pre-training \cite{mahajan2018exploring}, LGS forms a representative benchmark and a pre-training dataset for downstream tasks that see distributions close to e-commerce.

\begin{table*}
\centering
\begin{minipage}{.54\textwidth}
    \begin{small}
    \caption{The classification accuracy of models trained on LGS shows that the LGS end leaves are well-separable.}
    \scalebox{.9}[1.]{\begin{tabular}{l!{\vrule width 2pt}c|c|cc}
        \toprule
        & LGS-117               & LGS-117 \hfill            & LGS-710 \hfill \\
        \textbf{LGS Accuracy}   & \scriptsize{from scratch} & \scriptsize{IN-pretrained}  & \scriptsize{IN-pretrained}  \\
        &                       &                           & \scriptsize{(Top-1)} \hspace{5mm} \scriptsize{\textcolor{gray}{(Top-5)}} \\
        \midrule
        After linear probing & --       & 69.58 \% & 60.72 \% \hspace{2mm} \textcolor{gray}{81.16} \% \\
        After fine-tuning    & 97.89 \% & 98.16 \% & 77.27 \% \hspace{2mm} \textcolor{gray}{89.09} \% \\
        \bottomrule
    \end{tabular}}
    \end{small}
    \label{tbl:LGS_acc}
    \vspace{4mm}
\end{minipage}
\hfill
\begin{minipage}{.44\textwidth}
    \centering
    \caption{The reconstruction quality of the MAE models trained on LGS and ImageNet, evaluated on COCO. The symbol $\uparrow$ denotes ``higher is better'' while $\downarrow$ means ``lower is better''.}
    \begin{small}
    \scalebox{.9}[1.]{\begin{tabular}{l!{\vrule width 2pt}c|c}
        \toprule
        \textbf{Training Dataset} & \textbf{Inception} ($\uparrow$) & $\;\;$ \textbf{FID} ($\downarrow$) $\;\;$ \\
        \midrule
        ImageNet-1k & 9.2930 & 114.60 \\
        IN pretrain$\rightarrow$IN+LGS & 9.1906 & 115.48 \\
        LGS & 10.187 & 91.387 \\
        \bottomrule
    \end{tabular}}
    \end{small}
    \label{tab:mae_coco}
\end{minipage}
\end{table*}

\section{Experiments} \label{sec:experiments}

\subsection{Image Classification and Reconstruction} 
\label{subsec:exp_setup}

In this subsection, we use image classification and reconstruction tasks to characterize the distributional difference between LGS and ImageNet. We consider the distributions of images as well as the labels.

\subsubsection{ImageNet Classifiers Do Not Readily Generalize to E-commerce}

The existing literature has shown that carefully constructed images collected in a bias-controlled manner can elicit a significant performance degradation on classifiers trained on ImageNet \cite{Andrei2019ObjectNet}. By applying pre-trained ImageNet classification models to the LGS-Overlap dataset without further training, we show that such out-of-distribution examples naturally exist in the e-commerce domain. Specifically, we use publicly available weights of a ResNet-50 model and a ConvNeXT-Base model. The ResNet-50 achieves a 74\% average recall across the 176 overlapping synsets over the ImageNet images, but the number noticeably reduces to 46.43\% on LGS-Overlap. The ConvNeXT-Base obtains 79.00\% and 50.14\% on ImageNet and LGS-Overlap, respectively. This difference highlights that existing ImageNet models do not readily transfer to LGS instances. 
In addition to having a different label distribution, the e-commerce domain forms a natural distribution shift even for the classes that also exist in ImageNet. While taxonomy standardization techniques exist, aligning and merging the label space is still hard in general. Thus, a pre-training dataset that is more aligned with e-commerce is necessary, and LGS fulfills this role.

We further show that LGS end leaves are well-separable, verifying that the performance degradation of ImageNet models is caused by the distribution mismatch and not the ambiguity of the LGS classes. Note that \Cref{tbl:LGS_acc} illustrates that the models learned on LGS-117 / LGS-710 can achieve high accuracy on LGS-117 / LGS-710. Specifically, we consider the ``linear probing followed by fine-tuning'' training schedule, a transfer learning scheme that has been shown to improve the robustness against distribution shift by avoiding significant distortions of the pre-trained weights.

\subsubsection{Non-classification Visual Feature Extractors Can Generalize}

\begin{figure*}[!ht]
    \centering
    \includegraphics[width=\textwidth]{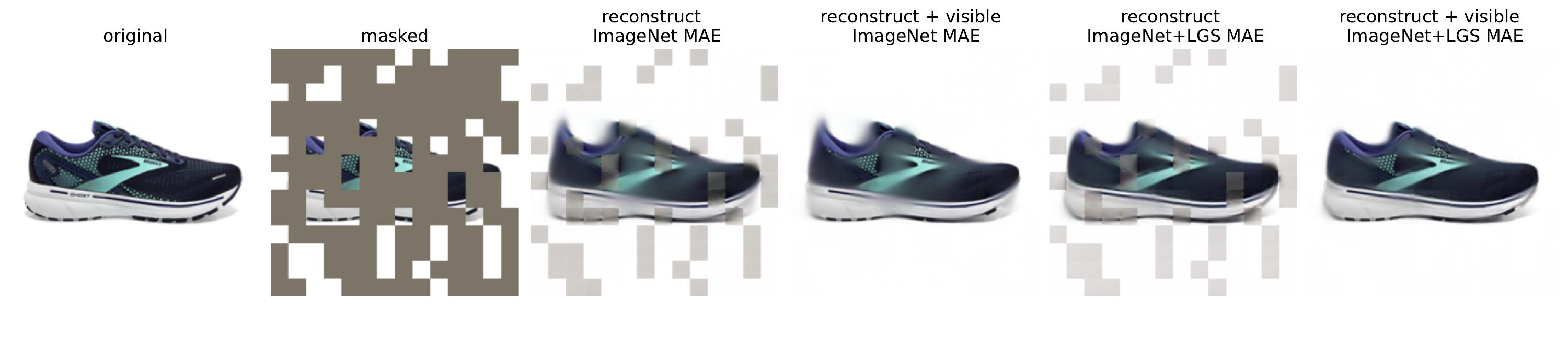}
    \vspace{-1cm}
    \caption{While an MAE trained on ImageNet can reasonably reconstruct an LGS image, adding LGS instances to the training improves the reconstruction quality.}
    \label{fig:MAE_on_LGS}
\end{figure*}

\begin{figure*}[!ht]
    \centering
    \includegraphics[width=\textwidth]{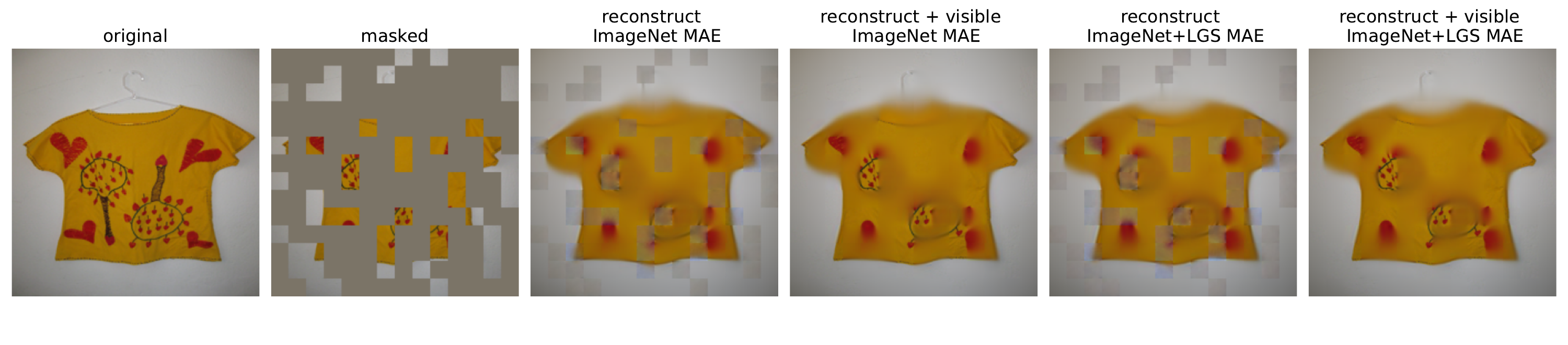}
    \vspace{-1cm}
    \caption{Adding LGS instances to the training also improves the reconstruction on some ImageNet instances.}
    \label{fig:MAE_on_IN}
\end{figure*}

Since the image-label correspondence is different between LGS and ImageNet, we use self-supervised training to isolate this mismatch and focus on the distribution of images. In the context of transfer learning, since self-supervised training does not use labels, it circumvents the issue of label space mismatch between target and source domains, which has been shown to undermine the quality of transfer learning. Masked AutoEncoder (MAE) \cite{He2022MAE} is a self-supervised method designed for pre-training. Thus, we compare the performance of an MAE trained on ImageNet only with an MAE trained on ImageNet and LGS-710. Figure \ref{fig:MAE_on_LGS} shows that the MAE trained on ImageNet can reconstruct a reasonable LGS image, but the reconstruction quality of the ImageNet+LGS model is better, demonstrating that LGS can be used to learn e-commerce visual features.

\begin{table}[!tb]
    \begin{center}
    \caption{Linear probing accuracy of the self-supervised MAE models with three different initializations. A: baseline ImageNet MAE model \cite{He2022MAE}, B: LGS MAE model, C: LGS+ImageNet MAE model. Specifically, C is initialized with A followed by 150 epochs on mixed Imagenet and LGS-710 data (1:1 ratio). Fine-tuning on LGS-117 and Imagenet datasets used 40 and 60 epochs, respectively.}
    \begin{small}
    \scalebox{.94}[1.]{\begin{tabular}{l!{\vrule width 2pt}c|c|c}
        \toprule
        & \multicolumn{3}{c}{\textbf{MAE Training Setting}} \\
        \textbf{Linear probing dataset} & \textbf{A} & \textbf{B} & \textbf{C} \\
        \midrule
        LGS-117 \scriptsize{(40 epochs)}     & 72.98 \% & 76.37 \% & 76.87 \% \\
        \hline
        ImageNet-1k \scriptsize{(60 epochs)} & 67.78 \% & 46.37 \% & 65.29 \% \\
        \bottomrule
    \end{tabular}}
    \end{small}
    \end{center}
    \label{tab:MAE_LP}
\end{table}

To quantitatively demonstrate the generalizability of the vision feature extractors, we evaluate the reconstruction performance of the MAE models trained on LGS and ImageNet on COCO. The qualities of the raw reconstructions obtained by the models are presented in \Cref{tab:mae_coco}. While LGS is more domain-specific compared with ImageNet and COCO (both of which cover a wide range of domains), the MAE trained on LGS is able to generate COCO images with higher qualities compared with the ImageNet model. Furthermore, we use \Cref{tab:MAE_LP} to show that upon the visual embeddings learned jointly on ImageNet and LGS, a linear classifier with a satisfactory performance can be learned on both ImageNet and LGS. The above results verify that the feature extractors can generalize between LGS and general-domain datasets, despite the separation of the intermediate visual embeddings (which are visualized in \Cref{sec:umap}).

\begin{table*}[!tb]
	\centering
	\caption{ImageNet$\to$LGS-710 two-phase pre-training improves downstream linear probing accuracy for downstream tasks including CIFAR-100, Fashion MNIST, and Clothing1M. On Clothing1M, whose data also comes from the e-commerce domain, the LGS-pre-trained features also improve end-to-end fine-tuning performance. For Clothing1M, we only use its clean training set, whereas Clothing1M (10\%) is a few-shot setup that trains on a 10\% subset of the clean training set.}
	\begin{small}
	\scalebox{.97}[1.]{\begin{tabular}{l!{\vrule width 2pt}c|c|c|c|c!{\vrule width 2pt}c|c}
	\toprule
	& \multicolumn{5}{c!{\vrule width 2pt}}{\textbf{Linear Probing}} & \multicolumn{2}{c}{\textbf{End-to-end training}} \\
	\textbf{Pre-training Setup} & CIFAR-10 & CIFAR-100 & \begin{tabular}[c]{@{}c@{}}Fashion \\ MNIST\end{tabular} & \begin{tabular}[c]{@{}c@{}}Clothing1M \\ (10 \%)\end{tabular} & \begin{tabular}[c]{@{}c@{}}Clothing1M\\ (100 \%)\end{tabular} & \begin{tabular}[c]{@{}c@{}}Clothing1M\\ (10 \%)\end{tabular} & \begin{tabular}[c]{@{}c@{}}Clothing1M \\ (100 \%)\end{tabular} \\
	\midrule
	ImageNet & 61.97 & 40.46 & 79.68 & 59.74 & 67.57 & 65.69 & 74.81 \\
	ImageNet$\to$LGS-117 & 59.83 & 35.57 & 80.39 & 64.48 & 69.67 & 68.16 & 75.47 \\
	ImageNet$\to$LGS-710 & 58.81 & 42.21 & 82.18 & 64.16 & 70.06 & 65.85 & 74.51 \\
	\bottomrule
	\end{tabular}}
	\end{small}
	\label{tab:LGS_pretrain}
\end{table*}

Based on the above observations, we infer that the e-commerce data distribution, represented by the LGS dataset, significantly differs from existing general datasets in the label space, while visual features can generalize. Thus, LGS is an ideal pre-training dataset for downstream tasks whose class distributions align with the e-commerce domain.

\subsubsection{LGS Supplements ImageNet as a Pre-training Dataset} \label{sec:LGS_pretrain}

LGS can also widen the span of the pre-training distribution when used in conjunction with ImageNet, acting as a bridge between general visual features and domain-specific applications. Specifically, \Cref{tab:LGS_pretrain} shows that a two-phase ImageNet$\to$LGS-710 weakly-supervised pre-training scheme produces features more suitable for fine-tuning on common downstream tasks. On e-commerce-related downstream datasets such as Clothing1M, the models pre-trained on LGS also excel in both linear probing and end-to-end settings. 

In linear probing experiments, we observe that incorporating in-domain pre-training~(both LGS-117 and LGS-710) results in better performance (2\% absolute) compared to ImageNet pre-training. Moreover, in limited-data settings, we observe less model regression compared to the full-data setups. For example, for fine-tuning a linear classifier on 10\% of the Clothing1M-clean dataset, the ImageNet pre-trained model regresses more (11.5\% relative) compared to LGS-117 and LGS-710 pre-trained models~(7.4 and 8.4\% relative respectively). 
When models are trained end-to-end, we observe that the pre-training setup is less critical in fine-tuning the full Clothing1M-clean training dataset. However, for limited-data experiments, filtering out under-represented classes (LGS-117) in pre-training helps with the downstream fine-tuning results~(2\% absolute) compared to both ImageNet and LGS-710 datasets. 

In \Cref{sec:gradcam} in the supplementary materials, we use GradCam \cite{GradCam, Selvaraju17} to visualize the representations learned by the classification models, demonstrating that the LGS models look for much more localized patterns that are relevant to e-commerce classification.

\subsection{Caption Generation} \label{sec:caption}

In this section, we illustrate that the distinct distribution of LGS benefits vision-language bi-modal tasks. Specifically, we study the efficacy of image-captioning (IC) models trained on traditional datasets in predicting LGS-type descriptions. We also evaluate the performance of LGS-trained models in generating attribute-rich image captions that would otherwise not be possible for models trained on more traditional datasets.

In this experiment, we utilize a bi-modal modeling framework based on OFA \cite{wang2022OFA}, a recently proposed encoder-decoder architecture that has achieved state-of-the-art performances in many language-vision tasks.
For each LGS image, the corresponding caption can be constructed by concatenating the ``product description'' strings in various orders. Specifically, we create three types of captions: 
\begin{enumerate}
    \setlength\itemsep{-.15em}
    \item LGS-title : title and brand name;
    \item LGS-taxonomy : product taxonomy;
    \item LGS-description: concatenated bullet strings.
\end{enumerate}

\begin{table}[!tb]
  \begin{center}
    \caption{IC model performance of image-captioning task evaluated on different combinations of training and evaluation datasets.}
    \begin{small}
    \begin{tabular}{l!{\vrule width 2pt}c!{\vrule width 2pt}c}
      \toprule
      \textbf{Training Set} & \textbf{Test Set} & \textbf{METEOR ($\uparrow$)} \\
      \midrule
      LGS-title & LGS-title         & 0.184 \\
      LGS-description & LGS-title   & 0.161 \\
      LGS-taxonomy & LGS-taxonomy   & 0.584 \\
      COCO & LGS-title              & 0.069 \\
      \bottomrule
    \end{tabular}
    \end{small}
    \label{tab:ofa_perf}
  \end{center}
\end{table}
The OFA IC model was trained on the three types of LGS inputs as well as on the traditional COCO dataset. The IC model performance in terms of its ability to predict the appropriate target string is tabulated in \Cref{tab:ofa_perf}.

\subsection{Text-to-Image Generation}

\begin{figure}[!tb]
    \centering
    \small{LGS Examples} \\
    \begin{subfigure}[b]{0.3\linewidth}
        \centering
        \scriptsize{a photo of terez leggings navy camo stripe hi-shine bump squad leggings, e-commerce}
        \vspace{0.6cm}
    \end{subfigure}\hfill
    \begin{subfigure}[b]{0.3\linewidth}
        \centering
        \includegraphics[width=\textwidth]{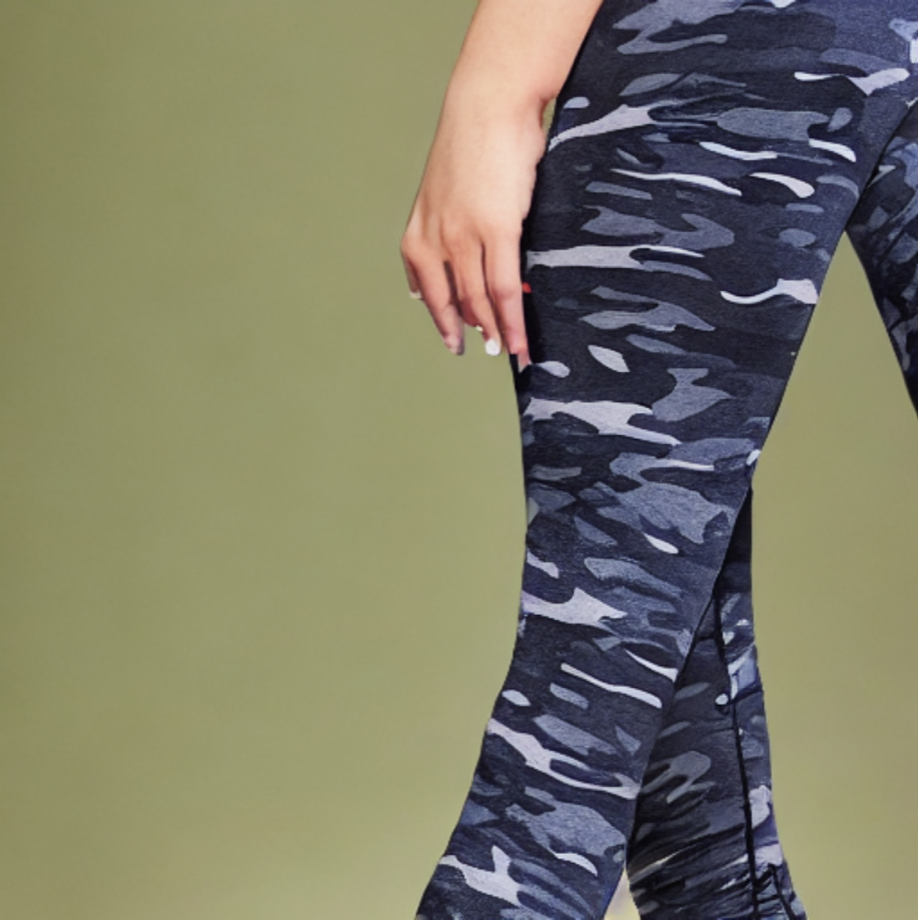}
    \end{subfigure}
    \begin{subfigure}[b]{0.3\linewidth}
        \centering
        \includegraphics[width=\textwidth]{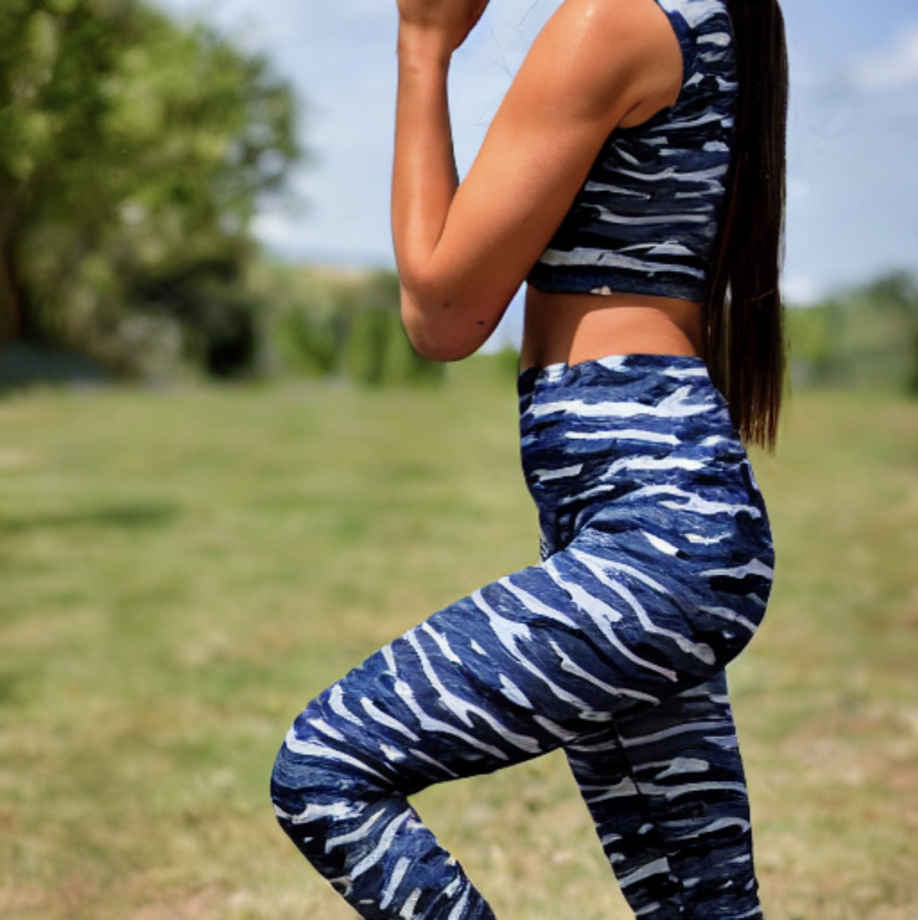}
    \end{subfigure}
    \begin{subfigure}[b]{0.3\linewidth}
        \centering
        \scriptsize{a photo of ana silver co. rings apatite ring size 8.25 (925 sterling silver) ring81016, e-commerce}
        \vspace{0.6cm}
    \end{subfigure}\hfill
    \begin{subfigure}[b]{0.3\linewidth}
        \centering
        \includegraphics[width=\textwidth]{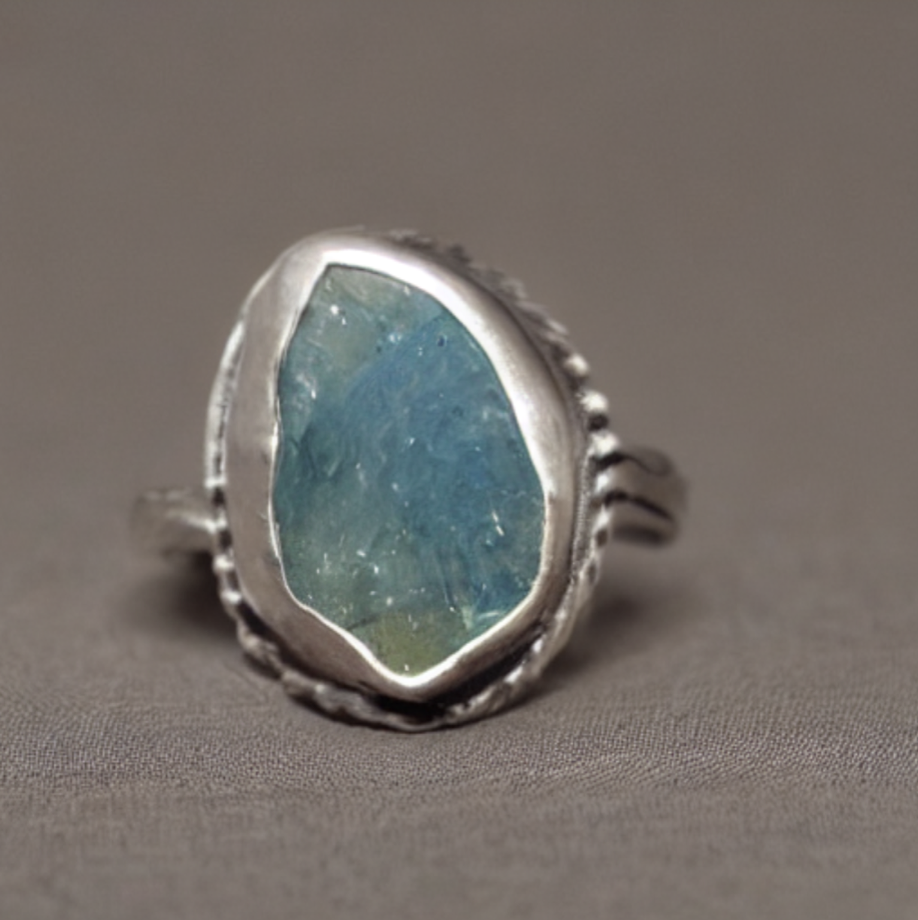}
    \end{subfigure}
    \begin{subfigure}[b]{0.3\linewidth}
        \centering
        \includegraphics[width=\textwidth]{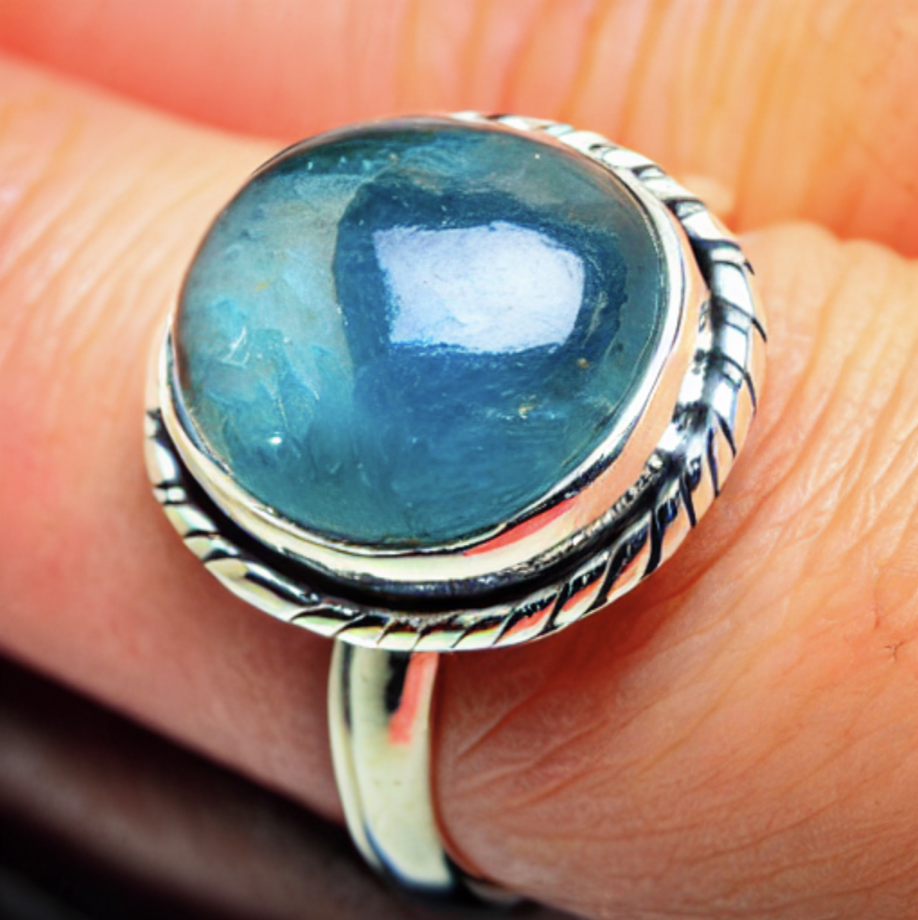}
    \end{subfigure}
    
    \vspace{0.1cm} \hrule \vspace{0.1cm}
    \small{DeepFashion InShop Examples} \\
    \begin{subfigure}[b]{0.3\linewidth}
        \centering
        \scriptsize{a photo of Dresses Ivory-navy Forever 21 Contemporary - Show the perfect amount of skin in this sleek, sophisticated surplice dress, e-commerce}
        \vspace{5mm}
    \end{subfigure}\hfill
    \begin{subfigure}[b]{0.3\linewidth}
        \centering
        \includegraphics[width=\textwidth]{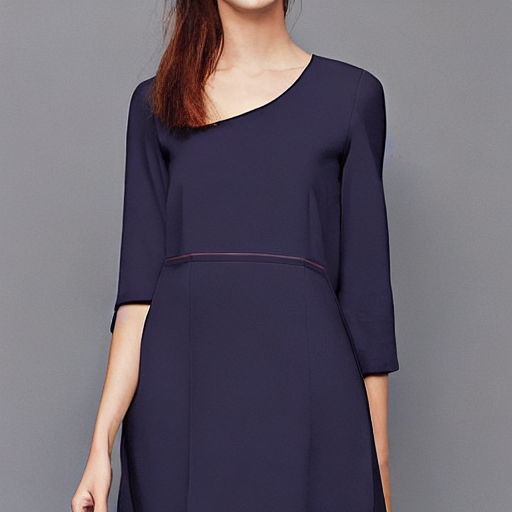}
    \end{subfigure}
    \begin{subfigure}[b]{0.3\linewidth}
        \centering
        \includegraphics[width=\textwidth]{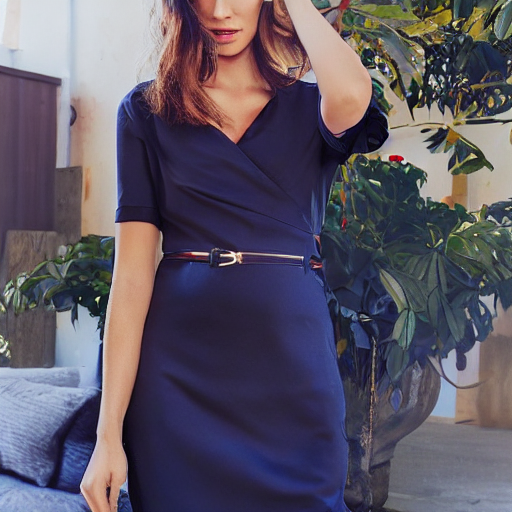}
    \end{subfigure}
    \begin{subfigure}[b]{0.3\linewidth}
        \centering
        \scriptsize{a photo of Pants Black-grey This jogger's easy, slouchy silhouette gets a little grit courtesy of its eye-popping print of photorealistic roses, e-commerce}
        \vspace{5mm}
        \caption{\scriptsize Input Prompt}
    \end{subfigure}\hfill
    \begin{subfigure}[b]{0.3\linewidth}
        \centering
        \includegraphics[width=\textwidth]{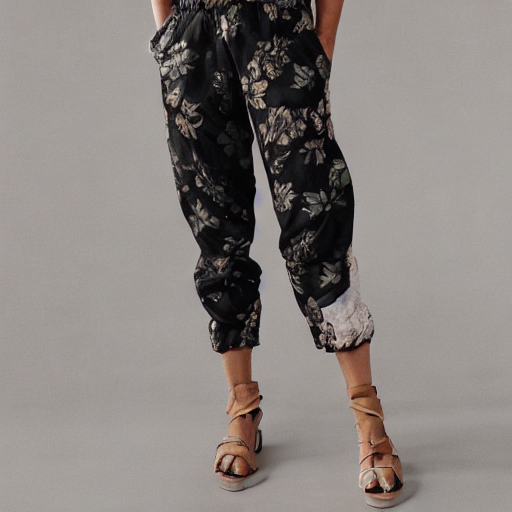}
        \caption{\scriptsize Vanilla SD}
    \end{subfigure}
    \begin{subfigure}[b]{0.3\linewidth}
        \centering
        \includegraphics[width=\textwidth]{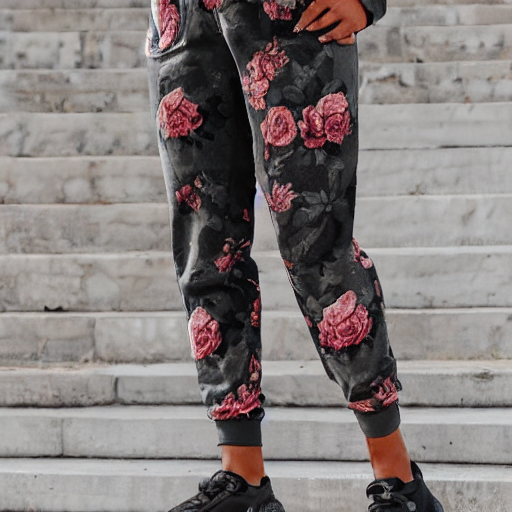}
        \caption{\scriptsize LGS-117}
    \end{subfigure}
    
    \caption{Qualitative comparisons of the generations of the Vanilla and the LGS-117-fine-tuned SD models in the general setting. The fine-tuned model generates more visually appealing images.}
    \label{fig:lgs_sd_general}
\end{figure}

Because of its high-quality e-commerce-focused images and bimodal nature, LGS is an ideal option for training text-to-image models in the e-commerce sector, serving as a bridge between general visual features and domain-specific applications. In this section, we use LGS to adapt the Stable Diffusion (SD) text-to-image generation method to two e-commerce scenarios: general and fine-grained. For both scenarios, we fine-tune based on the \texttt{sd-v1-4} (referred to as Vanilla) SD checkpoint.

\begin{table}[!t]
  \begin{center}
    \begin{small}
    \caption{Comparing the Vanilla SD and the LGS-117 fine-tuned model on LGS and DeepFashion datasets.}
    \label{tab:sd_perf}
    \begin{tabular}{l!{\vrule width 2pt}c|c}
      \toprule
      \textbf{Model} & \textbf{Test Set} & \textbf{FID ($\downarrow$)} \\
      \midrule
      Vanilla & LGS Val & 25.3498 \\
      Vanilla + LGS-117 & LGS Val & \textbf{24.1952} \\ \hline
      Vanilla & DeepFashion   & \textbf{62.9269} \\
      Vanilla + LGS-117 & DeepFashion   & 74.0185 \\
      \bottomrule
    \end{tabular}
    \end{small}
  \end{center}
\end{table}

For the general setting, we add a domain identifier to all training prompts associated with LGS images and guide the SD model to adapt to the e-commerce image style when this identifier is provided. The choice of the domain identifier is crucial, as the paper \cite{ruiz2022dreambooth} shows that a domain identifier with a strong prior should be avoided. For example, the word \texttt{retail} has a strong prior, and the pre-trained ``Vanilla'' SD model confidently associates it with (physical) retail stores. This behavior is undesirable for the goal of e-commerce style transfer. By analyzing the effects of various domain identifiers on the generations of the pre-trained SD model, we determine that the word ``e-commerce'' gives a weak prior and is a suitable identifier. We then construct the ground-truth training prompts for the LGS images in the format of \texttt{a photo of <brand> <end\_leaf> <title>, e-commerce}, where the \texttt{<end\_leaf>} refers to the end leaf of the taxonomy tree introduced in \Cref{sec:exp_classi}.
The ``Vanilla'' SD checkpoint is fine-tuned on one million LGS image-prompt pairs for 100k steps with a batch size of 24. Table \ref{tab:sd_perf} displays the quantitative results on an unseen validation set (5K image-prompt pairs) from LGS and a subset of the DeepFashion InShop dataset. The fine-tuning process enhances the performance of SD on LGS as expected. While the FID scores on DeepFashion are lower, the generations of the LGS-117 fine-tuned model are aesthetically more appealing. At this instant, there are no quantitative metrics that directly measure aesthetic superiority. Thus, we present \Cref{fig:lgs_sd_general} and the additional examples in \Cref{sec:text2img_supp} in the supplementary materials (Figures \ref{fig:t2i_sup} and \ref{fig:t2i_sup_inshop_imgs}) to demonstrate the aesthetic improvement qualitatively. The lower FID scores may indicate a distribution shift between LGS and DeepFashion images.

\begin{figure}
    \centering
    \begin{subfigure}[b]{0.3\linewidth}
        \centering
        \scriptsize{a photo of new balance men white running shoes}
        \vspace{0.9cm}
    \end{subfigure}\hfill
    \begin{subfigure}[b]{0.3\linewidth}
        \centering
        \includegraphics[width=\textwidth]{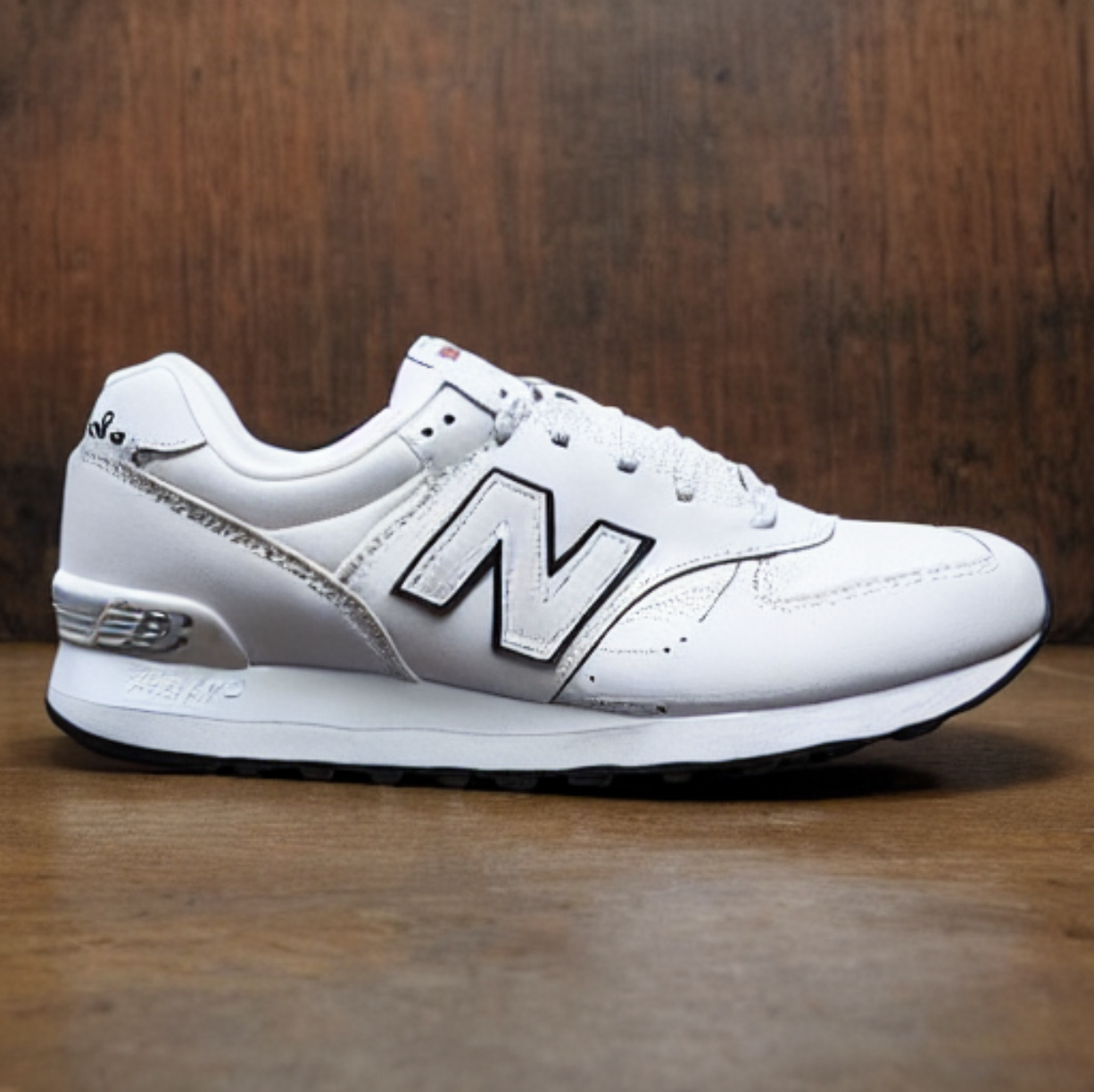}
    \end{subfigure}
    \begin{subfigure}[b]{0.3\linewidth}
        \centering
        \includegraphics[width=\textwidth]{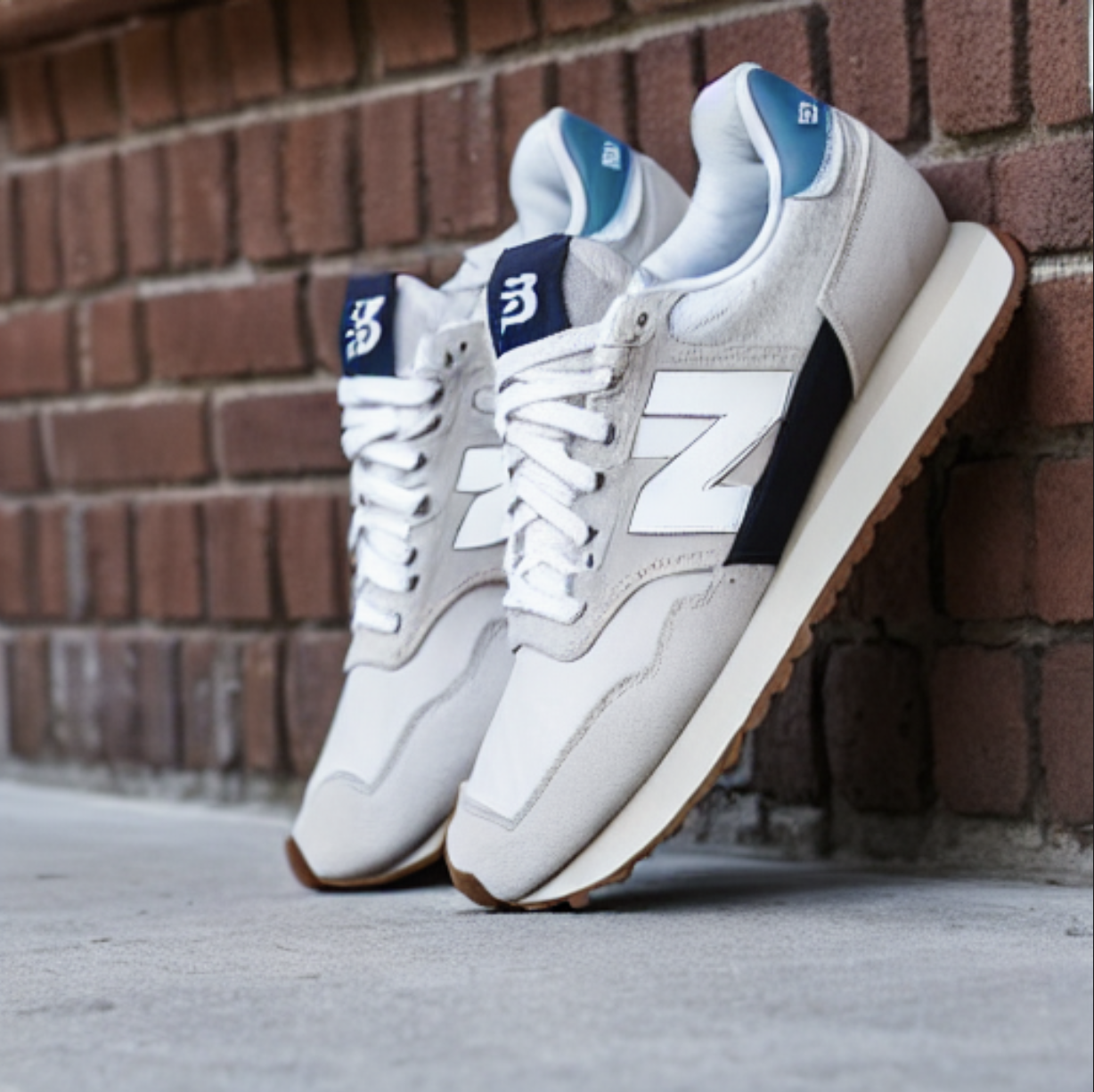}
    \end{subfigure}
    \begin{subfigure}[b]{0.3\linewidth}
        \centering
        \scriptsize{a photo of on running athletic shoes on running women's cloudswift road shoe 41.99578 in lake sky}
        \vspace{0.6cm}
    \end{subfigure}\hfill
    \begin{subfigure}[b]{0.3\linewidth}
        \centering
        \includegraphics[width=\textwidth]{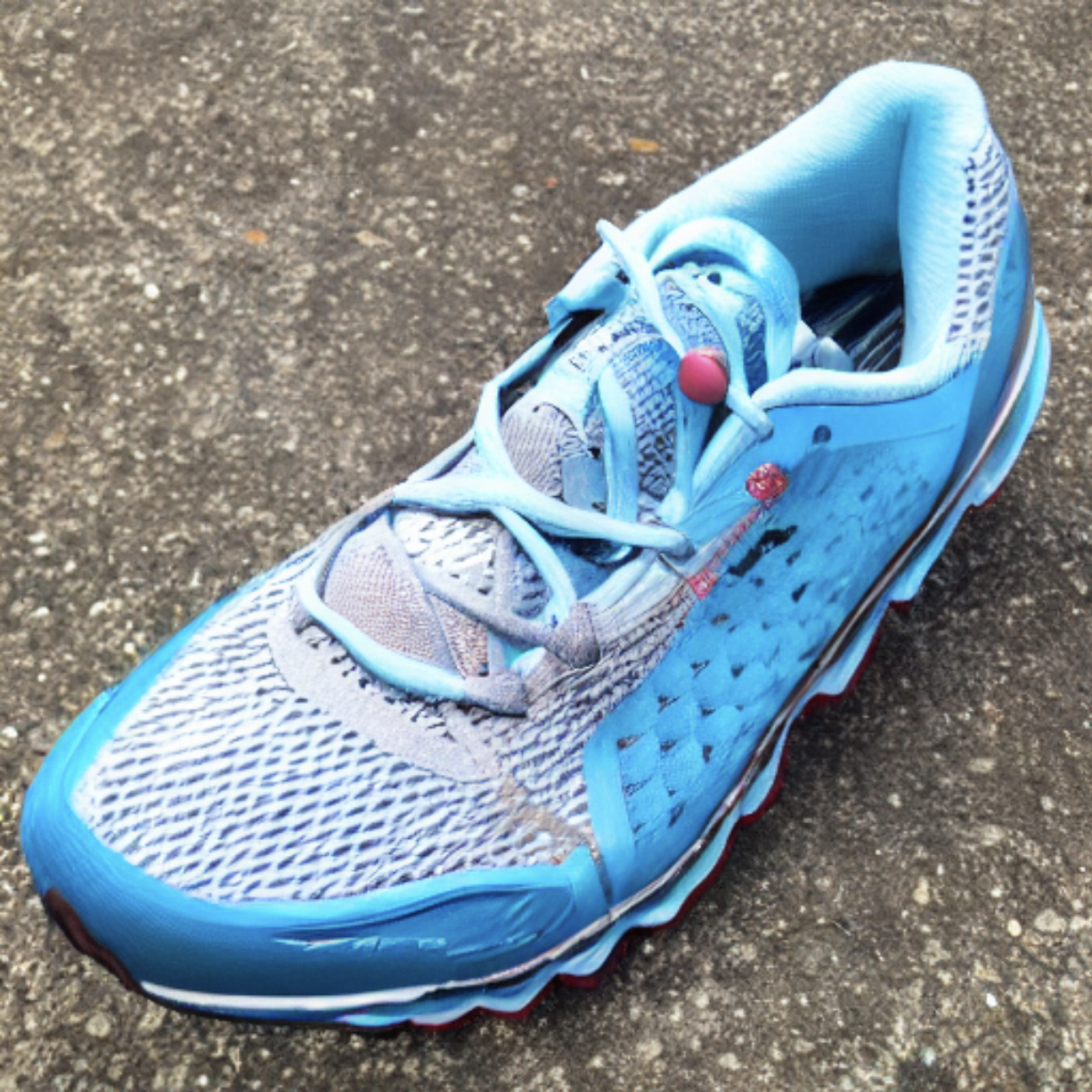}
    \end{subfigure}
    \begin{subfigure}[b]{0.3\linewidth}
        \centering
        \includegraphics[width=\textwidth]{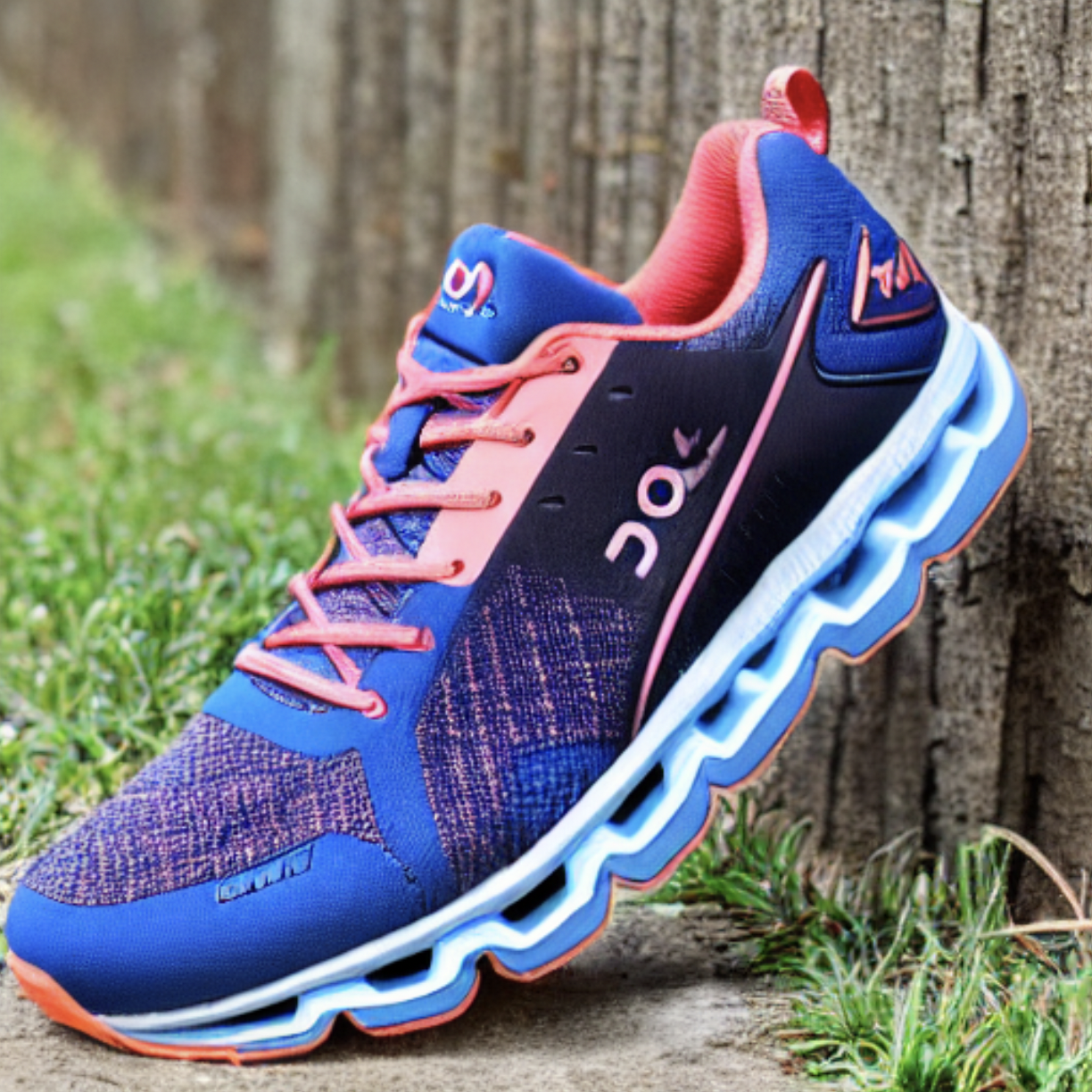}
    \end{subfigure}
    \caption{The LGS-117-fine-tuned SD model also generates more visually appealing images in the fine-grained setting. The prompts are from LGS.}
    \label{fig:lgs_sd_finegrain}
\end{figure}

For the fine-grained setting, we use data belonging to only a particular end leaf, using the same prompt without the additional identifier. The checkpoint is fine-tuned with 10k image-prompt pairs for 25k steps with a batch size of 6. We use the ``athletic shoes'' end leaf as an example and compare the generations before and after LGS-fine-tuning under the fine-grained setting in \Cref{fig:lgs_sd_finegrain}. As did the general setting results, the fine-grained examples also indicate that LGS helps adapt text-to-image models to e-commerce scenarios and improves image quality and aesthetics.

\section{Conclusion} \label{sec:discussion}

The Let's Go Shopping (LGS) dataset consists of 15 million pairs of publically-accessible diverse images and descriptive captions from e-commerce websites. Our efficient semi-automated gathering and annotation pipeline ensure scalable data collection. We then use LGS to show that while the categories associated with e-commerce data may not align with the general-domain pre-training datasets, visual feature extractors can be shared. Finally, we show that the distinct distribution offered by LGS and LGS's bi-modal nature can be beneficial for applications including image classification, image reconstruction, bi-modal representation learning, and text-to-image generation.

\section{Acknowledgment} \label{sec:acknowledgment}

We would like to appreciate Matteo Bruno, James Caulkins, Xiaowen Dong, Peter Grindrod, Jack Hessel, Sean Holloway, Dorothy Nicholas, Janet Pierrehumbert, and Julian Winkler for their valuable help and fruitful discussions.

We also appreciate Baillie Gifford, the Institute for New Economic Thinking at the Oxford Martin School, and the UK Engineering and Physical Science Research Council for funding our work at the University of Oxford.

This work was supported in part through the NYU IT High-Performance Computing resources, services, and staff expertise.

\newpage
{\small
\bibliographystyle{plain}
\bibliography{references}
}

\newpage
\onecolumn
\appendix

\section{Additional Analyses on LGS's Data Distribution}

\subsection{LGS End Leaf Histogram}

\renewcommand{\thefigure}{Supp-1}
\begin{figure*}
    \centering
	\includegraphics[width=\textwidth]{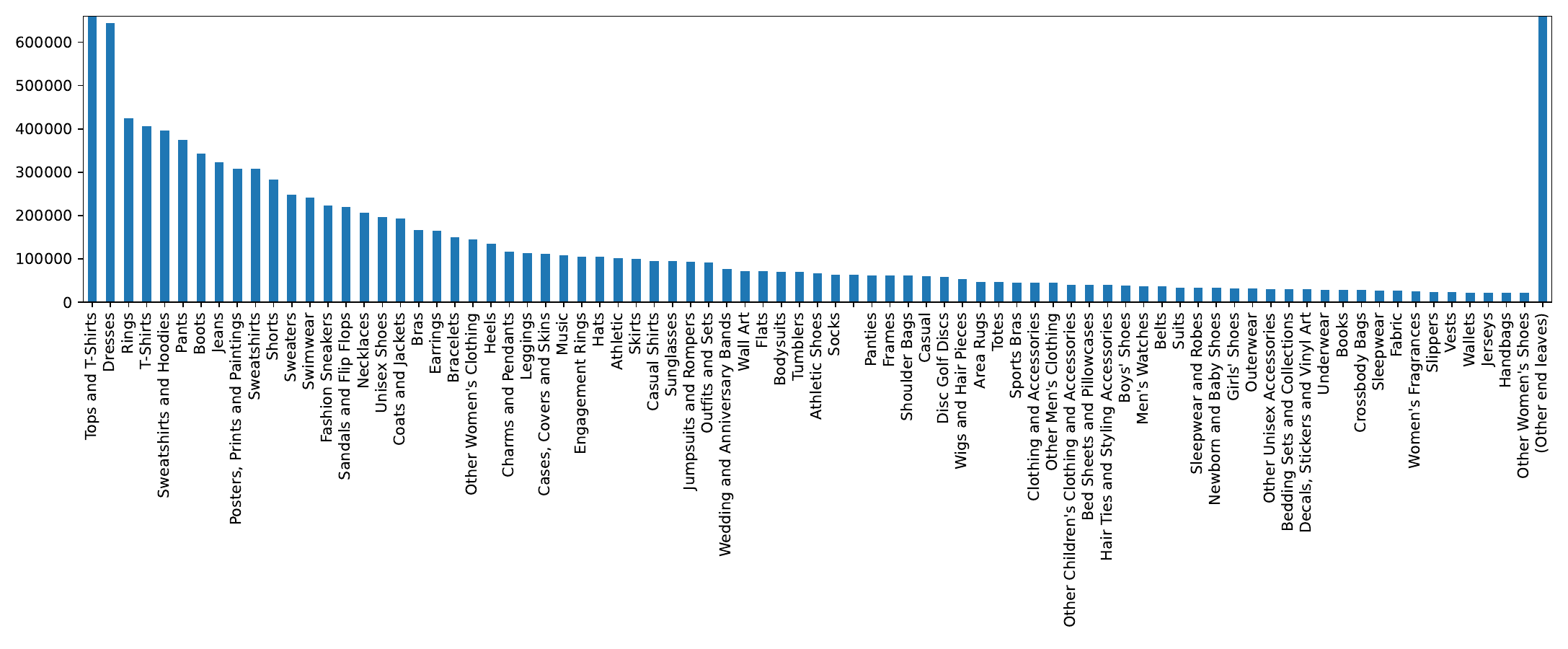}
	\vspace{-8mm}
	\caption{The instance counts of the 80 most popular LGS end leaves.}
	\label{fig:lgs_class_dist}
\end{figure*}

The instance counts of the LGS end leaves are displayed in \Cref{fig:lgs_class_dist}. The top 80 most popular end leaves encompass 83.28\% of the total instances, with the most popular \texttt{Tops and T-shirts} containing 16.23\% of the total instances.

\subsection{$n$-gram and POS Analysis of LGS Captions} \label{sec:ngrams}

\Cref{tab:ngrams} presents the comparisons of the uni-grams, bi-grams, and tri-grams of LGS. This comparison indicates that LGS is more linguistically diverse. The uni-grams and bi-grams of the two datasets are similar. However, we notice greater conceptual diversity for LGS within its tri-grams. Specifically, COCO's five most frequent tri-grams describe a group of objects and the relative position of the objects, whereas the LGS tri-grams encompass inherent properties of the commodities, including the size and the nature of each item.

In addition to the part-of-speech (POS) results presented in \Cref{sec:lgs_captions}, we use Figures \ref{fig:lgs_pos} and \ref{fig:coco_pos} to present the most common words per POS for LGS and COCO, respectively.

\renewcommand{\thefigure}{Supp-5}
\begin{figure*}[!tb]
	\centering
    \includegraphics[width=.42\textwidth, height=.27\textwidth]{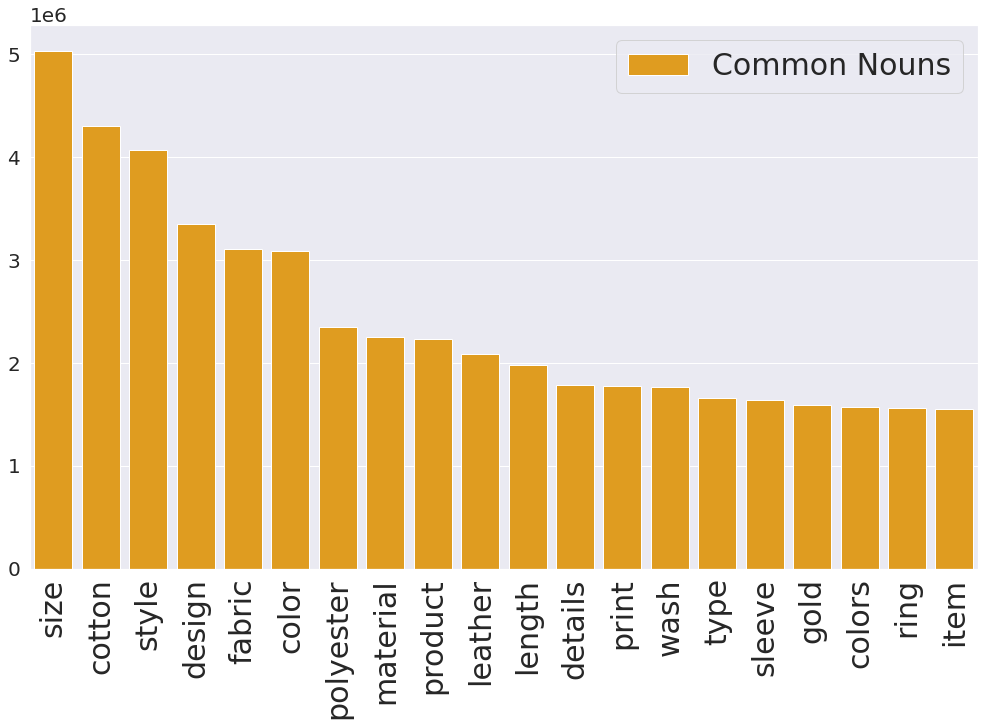}
    \includegraphics[width=.42\linewidth, height=.27\textwidth]{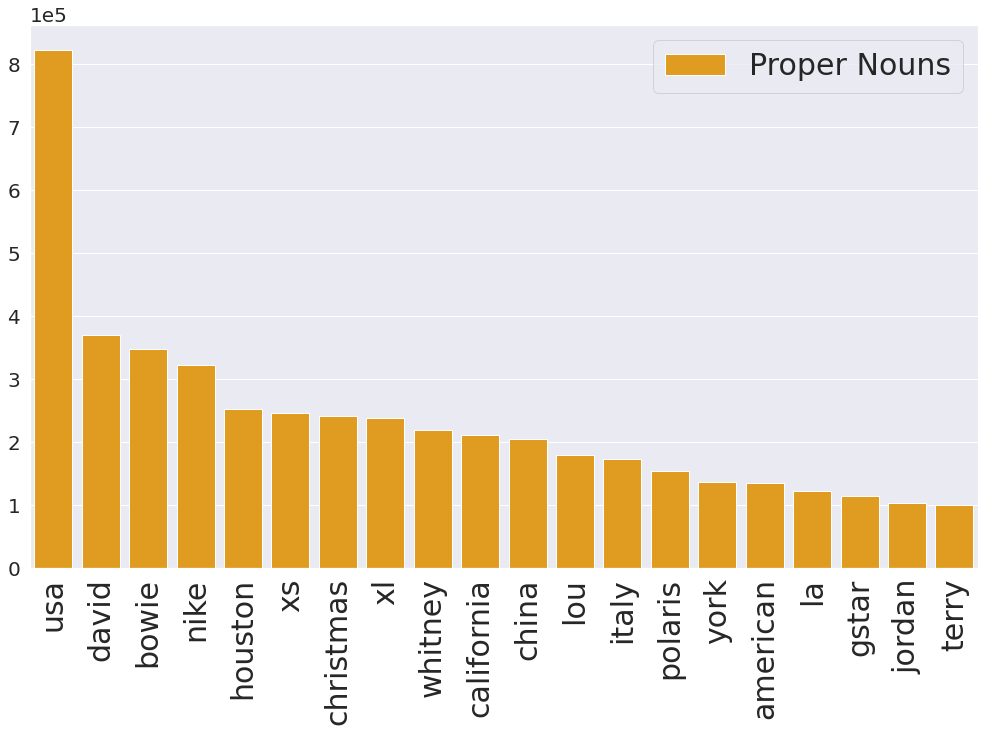} \\
    \includegraphics[width=.42\linewidth, height=.27\textwidth]{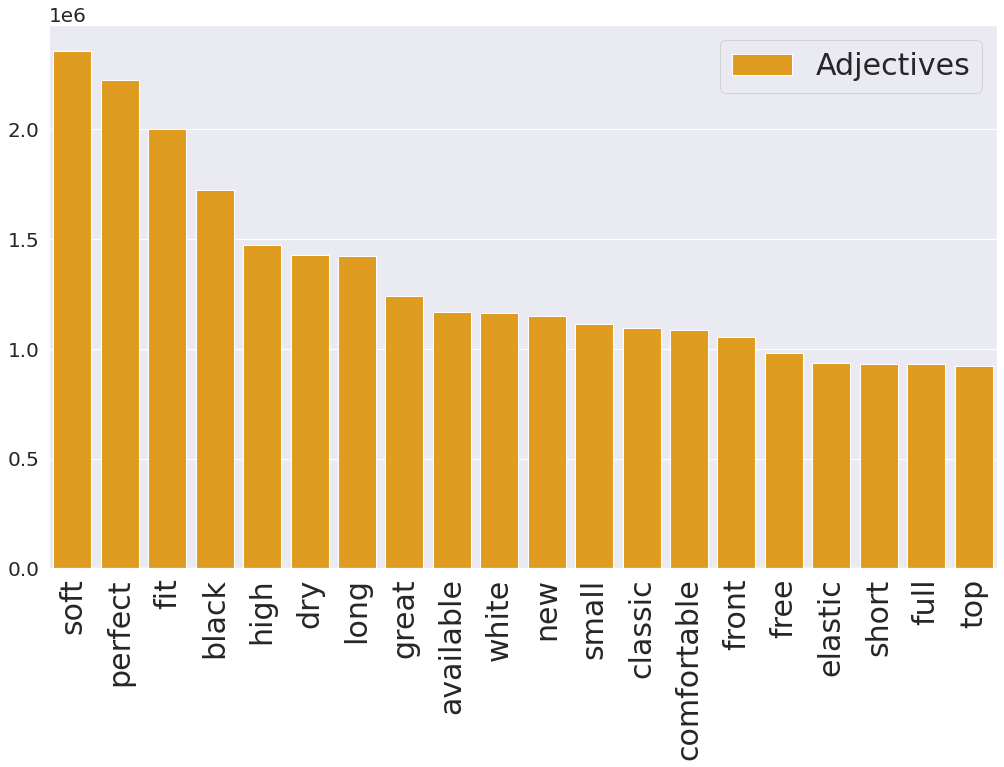}
    \includegraphics[width=.42\linewidth, height=.27\textwidth]{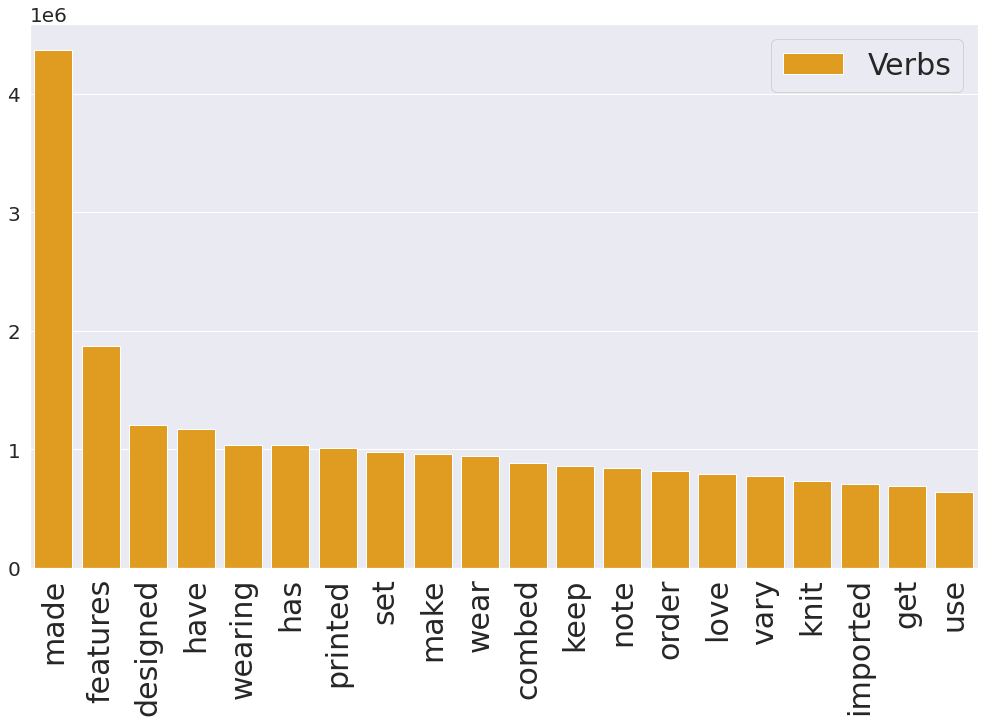}
    \vspace{-2mm}
    \caption{Top 20 most common words per POS for LGS.}
    \label{fig:lgs_pos}
\end{figure*}

\renewcommand{\thefigure}{Supp-6}
\begin{figure*}[!tb]
    \captionsetup[subfigure]{justification=Centering}
    \centering
    \includegraphics[width=.42\textwidth, height=.27\textwidth]{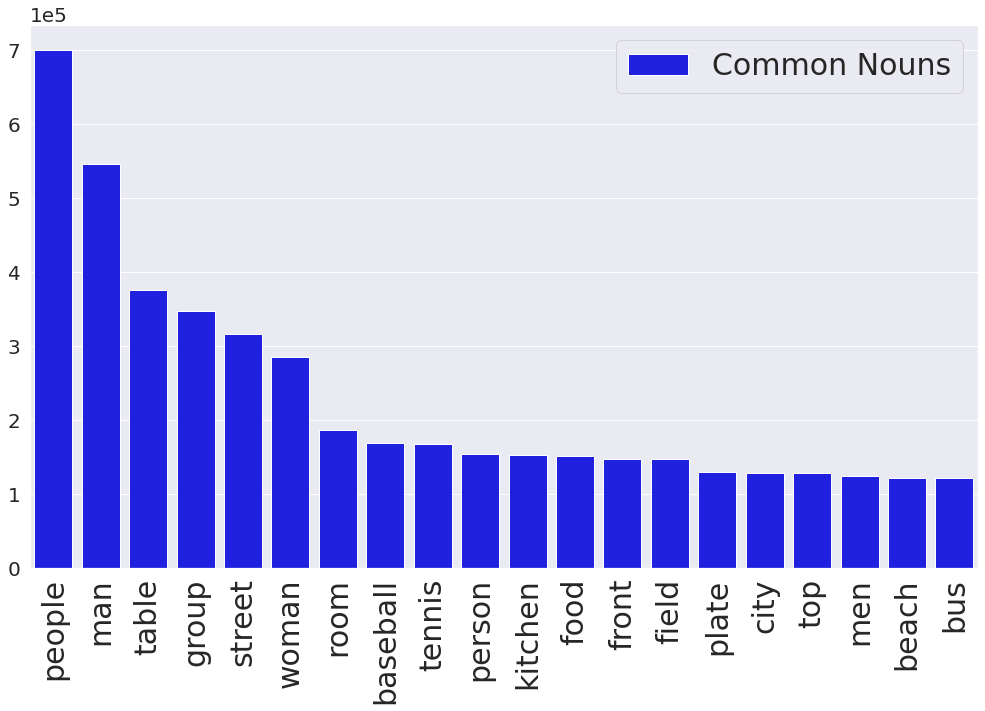}
    \includegraphics[width=.42\linewidth, height=.27\textwidth]{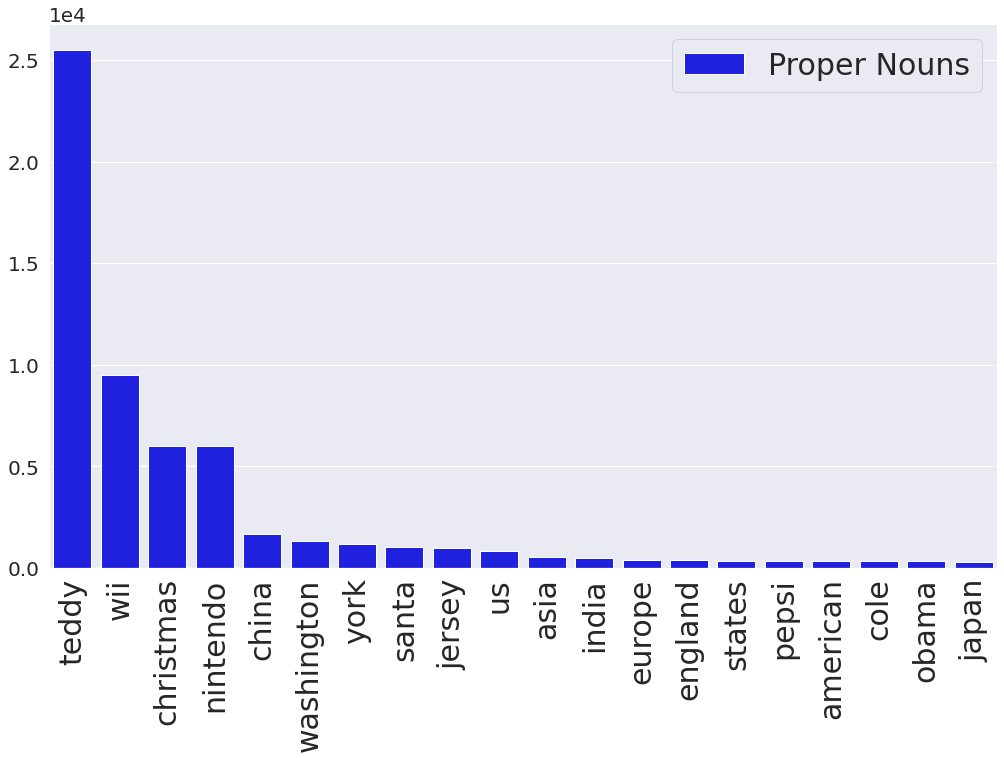} \\
    \includegraphics[width=.42\linewidth, height=.27\textwidth]{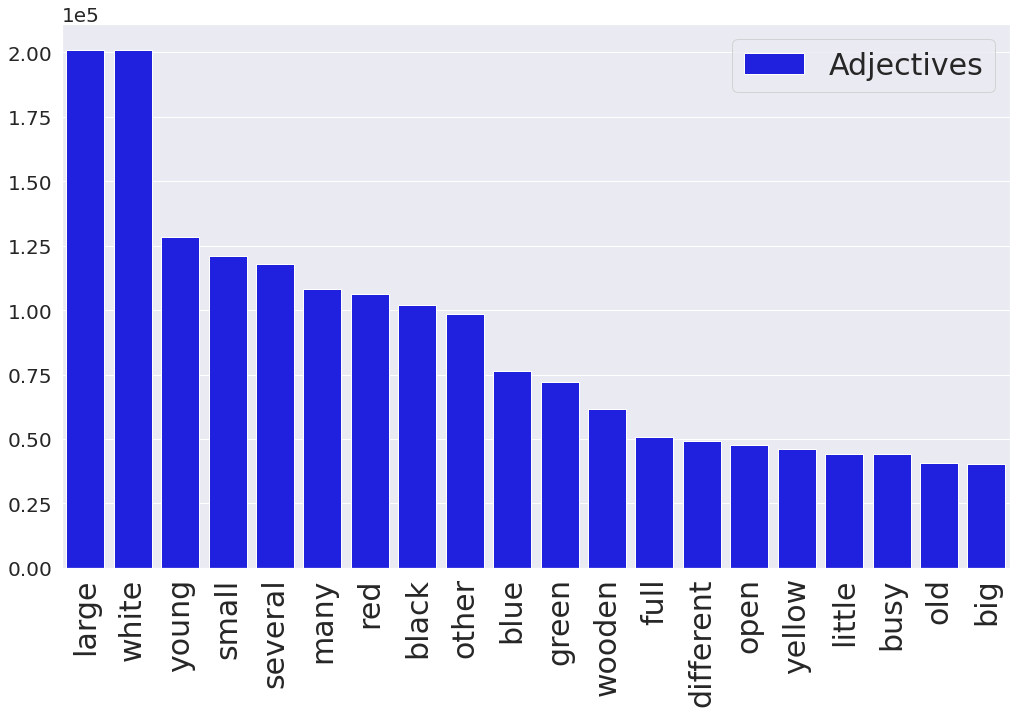}
    \includegraphics[width=.42\linewidth, height=.27\textwidth]{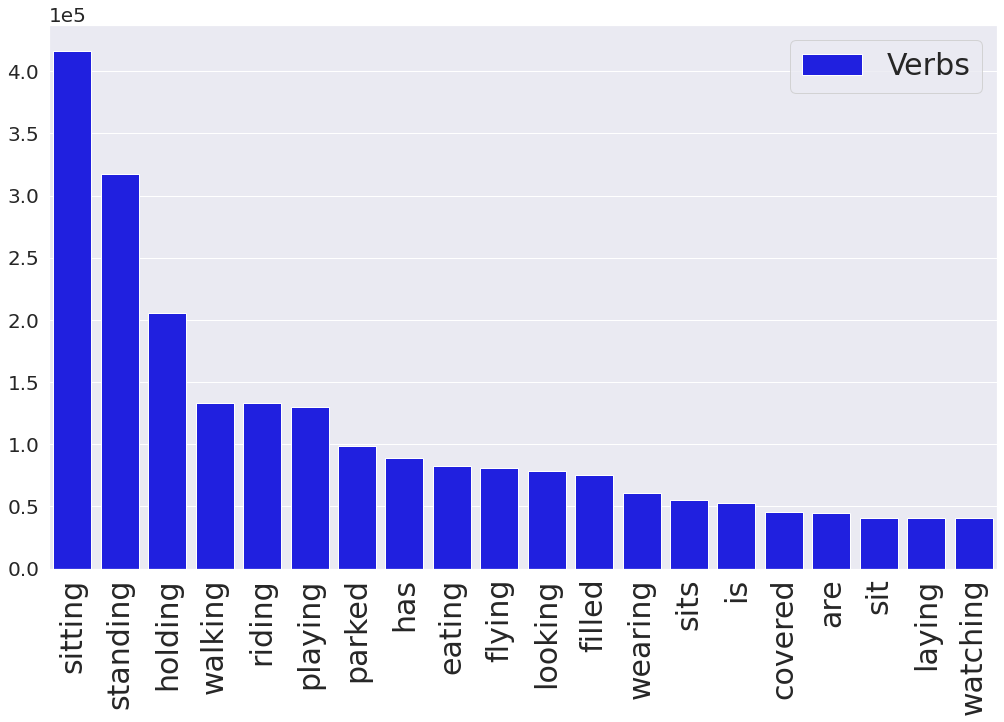}
    \vspace{-2mm}
    \caption{Top 20 most common words per POS for COCO.}
    \label{fig:coco_pos}
\end{figure*}

\renewcommand{\thetable}{Supp-1}
\begin{table*}[!tb]
  	\centering
    \caption{Comparing the $n$-gram statistics of LGS with that of COCO.}
    \begin{small}
    \begin{tabular}{l !{\vrule width 2pt} c | c !{\vrule width 2pt} c | c}
      \toprule
      & \multicolumn{2}{c !{\vrule width 2pt}}{\scalebox{.83}[1]{\textbf{Number of $n$-grams with occurrence $\geq 10$}}} & \multicolumn{2}{c}{\textbf{Five most frequent $n$-grams ($n = 1, 2, 3$)}} \\
      & $\qquad\quad$ \textbf{LGS} $\qquad\quad$ & \textbf{COCO} & \textbf{LGS} & \textbf{COCO} \\
      \midrule
      \textbf{uni-grams} & 364,802 & 17,009 & and, the, a, to, with & a, of, on, the, i \\
            \hline
      \textbf{bi-grams} & 4,054,418 & 184,882 &
      \scalebox{.98}[1]{with a, in the, of the, is a, for a} &
      \scalebox{.88}[1]{on a, in a, a man, of a, with a} \\
            \hline
     \multirow{2}*{\textbf{tri-grams}} & \multirow{2}*{8,900,084} & \multirow{2}*{462,653} &
     \scalebox{.86}[1]{true to size, made to order, this is a,} &
     \scalebox{.95}[1]{a group of, group of people} \\
     & & & \scalebox{.95}[1]{this item is, machine wash cold} &
     \scalebox{.85}[1]{in front of, next to a, on top of} \\
      \bottomrule
    \end{tabular}
    \end{small}
    \label{tab:ngrams}
\end{table*}

\newpage
\newpage
\section{Additional Analyses on LGS-trained Classifiers}

\subsection{How Features learned on ImageNet and LGS Differ} \label{sec:umap}

To understand how vision models interpret the ImageNet and LGS instances, we use a ResNet50 model sequentially trained on ImageNet and LGS-117 as the feature extractor, and use UMAP \cite{mcinnes2018umap} to visualize the high-dimensional ImageNet and LGS features in 2D figures. As shown in \Cref{fig:UMAP}, the ImageNet features form a cluster, while the LGS features form a less concentrated cluster. The separation of the two clusters is especially prominent at the first two layers.

\renewcommand{\thefigure}{Supp-2}
\begin{figure*}
    \centering
    \includegraphics[width=.28\textwidth]{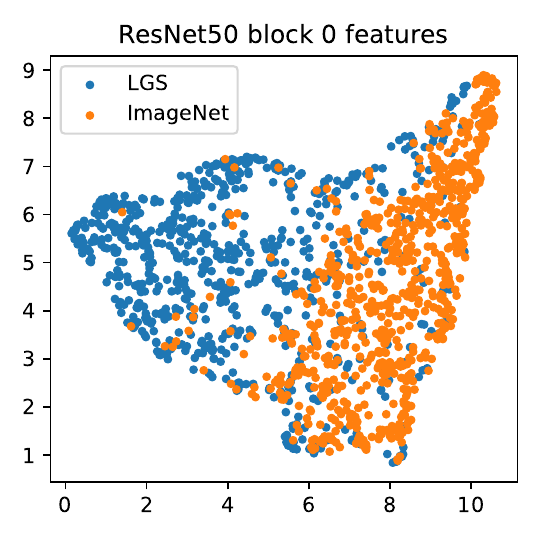}
    \includegraphics[width=.28\textwidth]{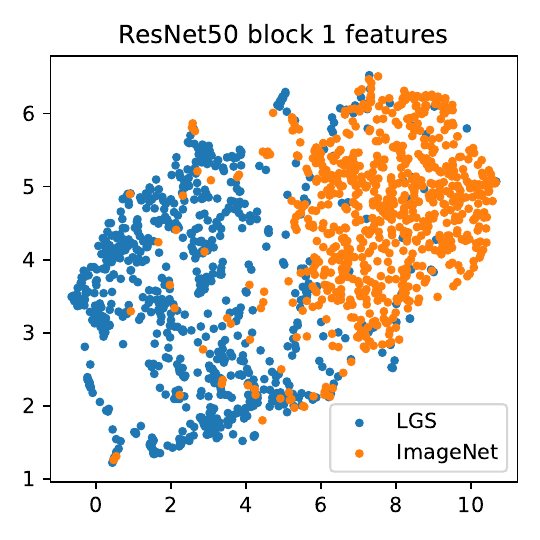}
    \includegraphics[width=.28\textwidth]{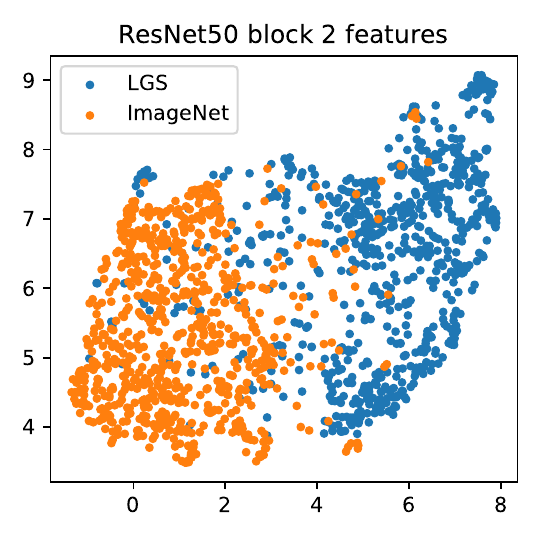} \\[-2mm]
    \includegraphics[width=.28\textwidth]{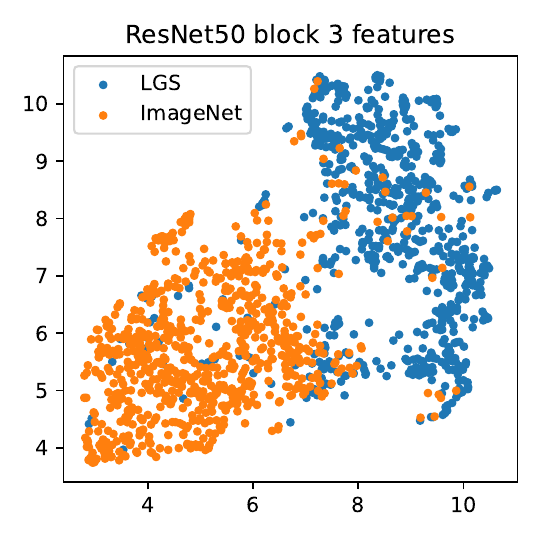}
    \includegraphics[width=.28\textwidth]{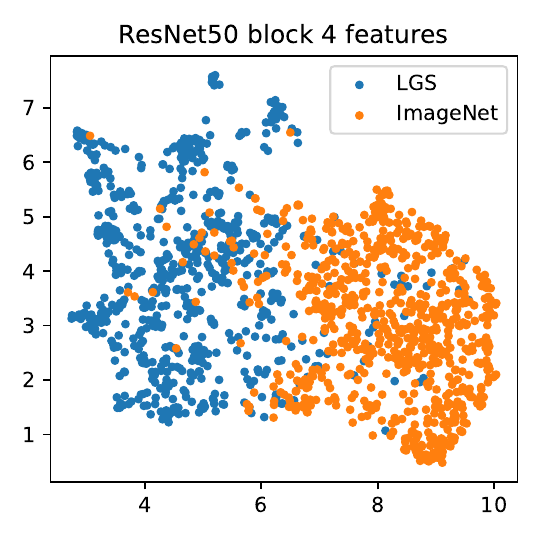}
    \vspace{-1mm}
    \caption{UMAP visualization of the ImageNet and LGS features extracted on a ResNet50 model trained on ImageNet and LGS.}
    \label{fig:UMAP}
\end{figure*}

\renewcommand{\thefigure}{Supp-3}
\begin{figure*}
    \centering
    \begin{subfigure}{.46\textwidth}
        \includegraphics[width=\textwidth, trim={1mm, 102mm, 1mm, 102mm}, clip]{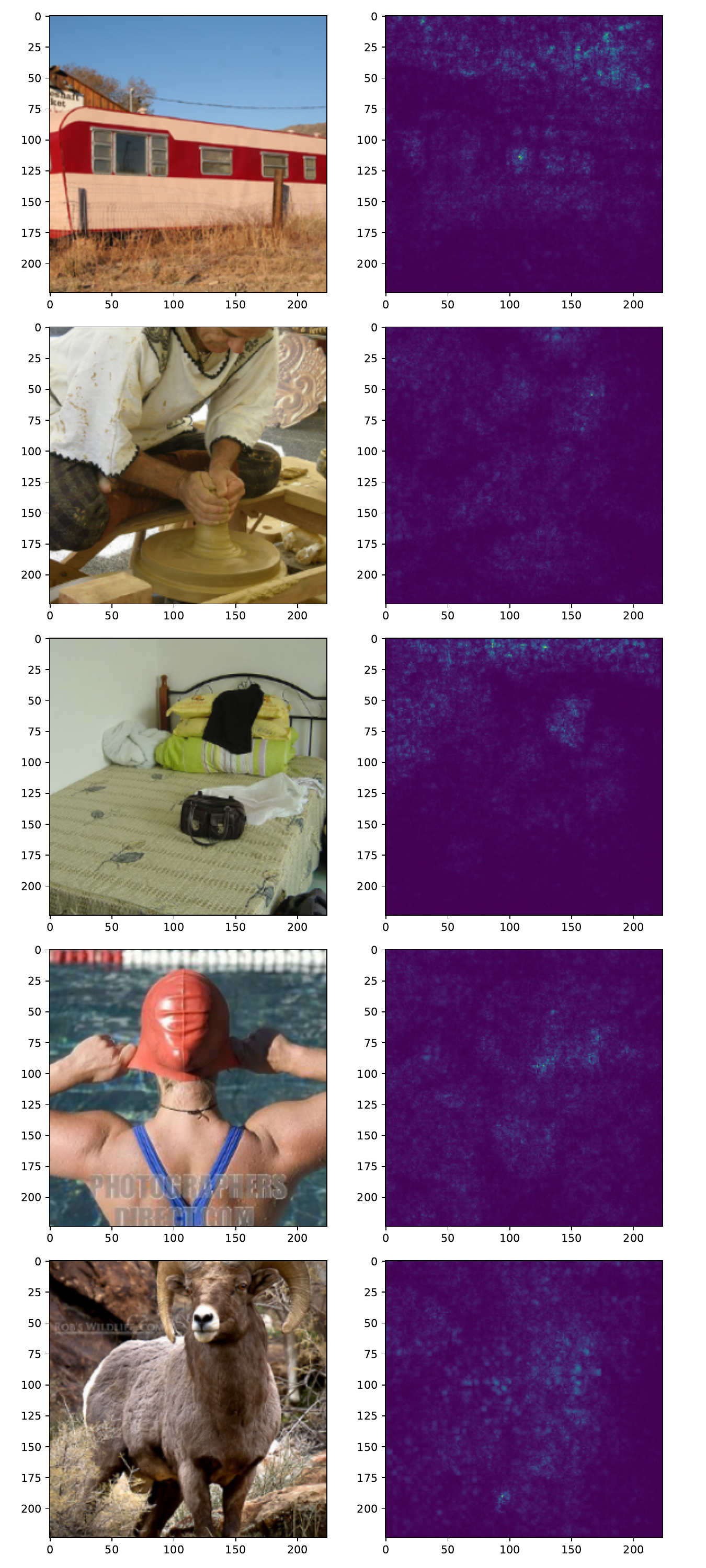}
        \caption{ImageNet examples.}
    \end{subfigure}
    \begin{subfigure}{.46\textwidth}
        \includegraphics[width=\textwidth, trim={1mm, 204mm, 1mm, 0mm}, clip]{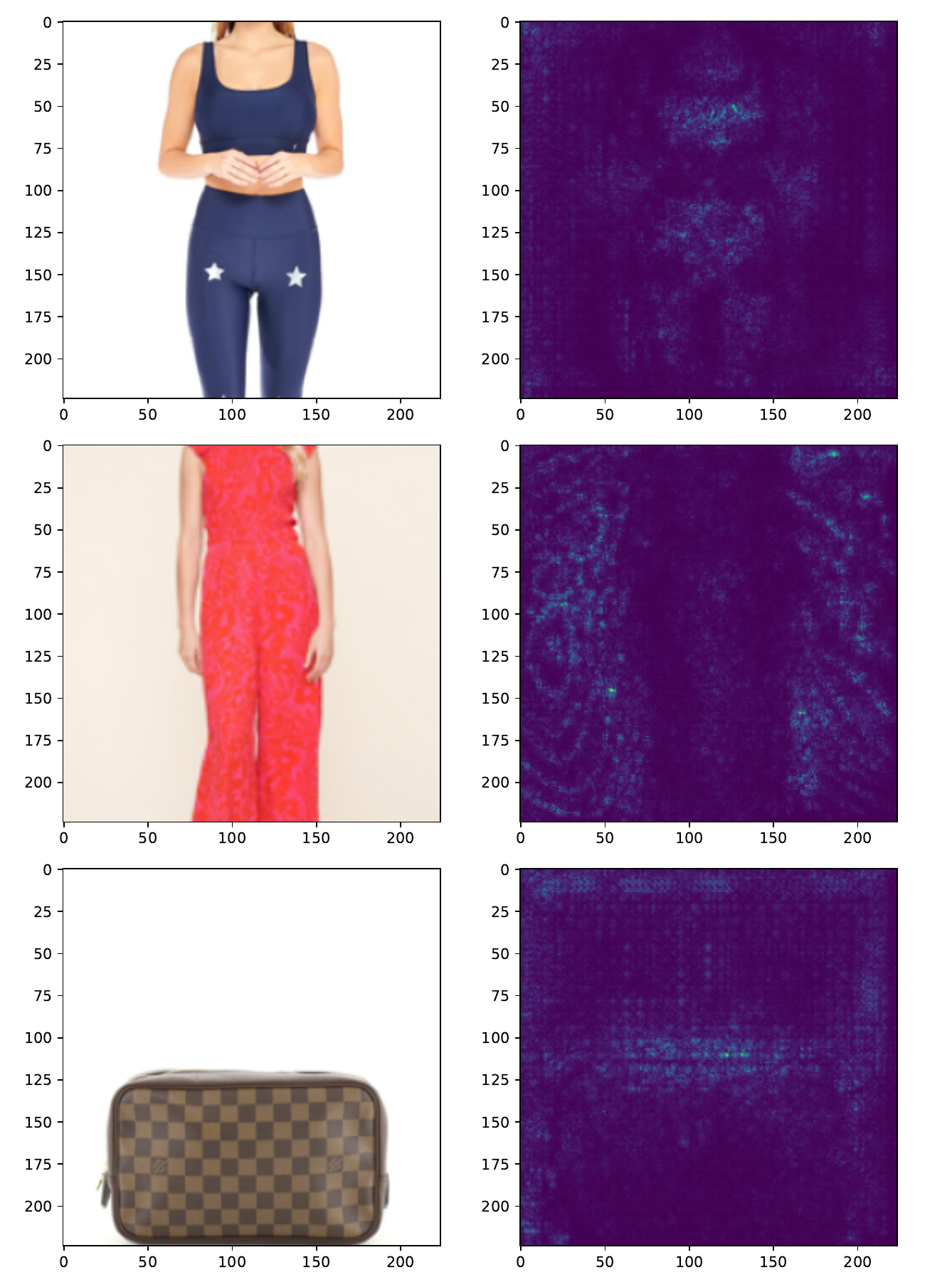}
        \caption{LGS examples.}
    \end{subfigure}
    \caption{The saliency map of the LGS-ImageNet binary classifier.}
    \label{fig:saliency}
\end{figure*}

As discussed in the main portion of the paper, many LGS product thumbnails consist of isolated foreground objects and clear backgrounds, while ImageNet instances are mostly natural images where the foreground blends into the background. Thus, we question whether the feature clustering is a consequence of this difference. To this end, we learn a binary classification linear header that predicts between LGS and ImageNet images based on the features extracted by the ResNet-50 model. We then visualize the saliency map of this binary model in \Cref{fig:saliency}. While the background is the most prominent difference between ImageNet and LGS to human eyes, the saliency maps demonstrate that the deep models look for more sophisticated patterns, which can vary across different images. Specifically, the foreground is emphasized in the first LGS example, while the background is more important in the second LGS instance. This observation aligns with the findings of \cite{ilyas2019adversarial}, which states that deep neural networks are not always understandable by humans.

\subsection{LGS Classification Models Look for Localized Patterns} \label{sec:gradcam}

\renewcommand{\thefigure}{Supp-4}
\begin{figure*}[!tb]
    \centering
    \includegraphics[width=.75\textwidth]{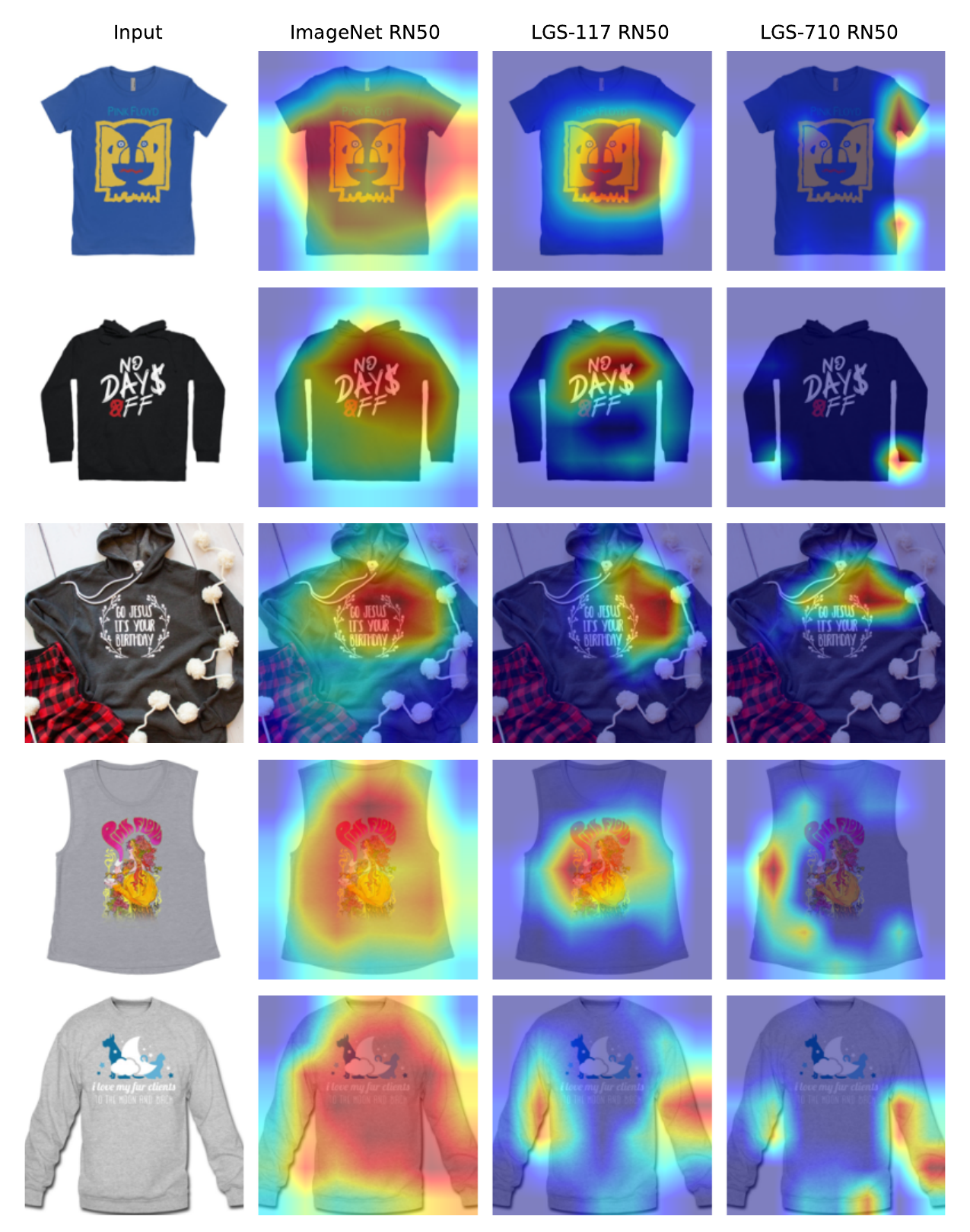}
    \caption{GradCam visualizations show that LGS classification models look for much more localized patterns.}
    \label{fig:gradcam}
\end{figure*}

In \Cref{fig:gradcam}, we use GradCam \cite{GradCam, Selvaraju17}, a framework that visualizes gradient activations of the input images, to demonstrate that the models trained on LGS look for much more localized patterns. Here, we draw examples from the ``sweatshirt'' synset in the LGS-Overlap dataset, and feed them into the three ResNet-50 models learned on ImageNet, LGS-117, and LGS-710, respectively. The gradient activation of the ImageNet model spreads across the entire image, while the LGS models return more concentrated gradient maps. Note that the gradient spikes produced by the LGS models mostly locate around the sleeves and the waist portion of the clothes. This makes sense because the LGS models are trained to differentiate various kinds of clothing-dominated e-commercial products. The portions highlighted by the LGS model gradient maps precisely correspond to the places where various types of clothes differ. For example, checking the sleeve length may be one of the easiest ways of distinguishing T-shirts from sweatshirts. Since the LGS-710 model was trained to classify more fine-grained types of products, it looks for even more localized patterns compared with the LGS-117 model.

\subsection{Linear Probing Details}

In this section, we discuss the implementation details for the linear probing experiments in \Cref{sec:LGS_pretrain}. In the existing literature, when ResNets (designed for $224 \times 224$ inputs) are adopted for tasks that use smaller input sizes, the first $7 \times 7$ convolution layer is often replaced with a $3 \times 3$ layer. We adopt this replacement for CIFAR and Fashion MNIST. During linear probing, we thus allow this modified, randomly reinitialized first layer to be optimized along with the output layer.

In \Cref{sec:LGS_pretrain}, we presented the improved linear probing results on CIFAR-100 and Fashion MNIST. We would like to highlight that linear probing is a practical training method, because when the batch normalization (BN) layers are jointly optimized alongside the first and the last layer, this modified ``linear'' probing scheme can achieve a performance that is comparable to end-to-end training \cite{Xu22}. Specifically, with learnable BN, a ResNet-50 model pre-trained on ImageNet$\to$LGS-710$\to$ImageNet achieves an accuracy of 71.41\% on CIFAR-100, compared with 69.47\% for an ImageNet-only model.

\newpage
$ $

\newpage
\section{Additional Text-to-Image Generation Discussions}

\subsection{Determining the Prompts for Text-to-Image Generation} \label{sec:text2img_supp}

Ensuring the quality of the input prompts is paramount for text-to-image models to generate realistic images. Our goal is to choose a prompt which generates images faithful to the metadata, performs relatively well in terms of Frechet Inception Distance (FID) score, and generalizes across datasets. 

To that end, we randomly selected 5,000 examples each from the LGS and DeepFashion InShop datasets. It is important to note that, for prompt engineering, the ground-truth images used for FID calculation are upscaled from $256 \times 256$, and the denoising diffusion implicit model steps (\texttt{ddim\_steps}) were lowered to 50 for inference. This resulted in lower scores than the experiment results (Table~\ref{tab:sd_perf}). However, the numbers are still indicative of relative performance.

Quantitatively, Prompts 3 and 4 perform significantly better on LGS, perform comparably on DeepFashion, and generalize well (Table~\ref{tab:t2i_sup_sd_results_all_prompts}).
Prompt 3 achieves better FID scores using the Vanilla model and performs slightly better on LGS.
Qualitatively, however, Prompt 4 generations are consistently better and more faithful to the metadata (Figure~\ref{fig:t2i_sup_prompt_eng_imgs}).
Therefore, we select Prompt 4 for our experiments.
This also reaffirms that these objective metrics are not strong indicators of the subjective aesthetic quality in this particular case, and should only be used as a loose relative measure. 
Figures~\ref{fig:t2i_sup} and \ref{fig:t2i_sup_inshop_imgs} show additional examples from the two datasets generated with Prompt 4.

\renewcommand{\thetable}{Supp-2}
\begin{table}[!t]
	\begin{center}
    \caption{The FID scores across prompts using a subset ($n = 5000$) of the LGS and DeepFashion InShop datasets.}
	\begin{small}
    \begin{tabular}
        {c!{\vrule width 2pt}c|c|c}
        \toprule
        \multirow{2}{*}{\textbf{Prompt ID}} & \multirow{2}{*}{\textbf{Model}} &  \multicolumn{2}{c}{\textbf{FID ($\downarrow$)}}
		\\
       	\cline{3-4}
       	& &  \textbf{LGS} & DeepFashion
       	\\
      	\midrule
      	1 & Vanilla &  \textbf{40.4437}  & \textbf{61.8519}
      	\\
        & Vanilla + LGS-117 & 42.7328 & 74.4327
        \\
      	\hline
      	2 & Vanilla & 42.1081  & \textbf{63.2344}
      	\\
        & Vanilla + LGS-117 & \textbf{42.0529}	& 77.7190
        \\
       \hline
      	3 & Vanilla & 36.7157 & \textbf{58.2189}
      	\\
        & Vanilla + LGS-117 & \textbf{36.1946}  & 79.3607
        \\
       	\hline
      	4 & Vanilla &  38.4101  & \textbf{62.9269}
      	\\
        & Vanilla + LGS-117 & \textbf{38.4100}  & 74.0185
        \\
      	\bottomrule
    \end{tabular}
    \end{small}
    \label{tab:t2i_sup_sd_results_all_prompts}
  \end{center}
\end{table}

\renewcommand{\thefigure}{Supp-7}
\begin{figure*}[!tb]
    \begin{center}
    \begin{subfigure}[b]{0.4\linewidth}
        \scriptsize
        \textbf{end\_leaf}: Jackets Vests
        \textbf{gender\_category}: Men
        \textbf{color}: Khaki
        \textbf{first sentence of description}:Made in a cotton-nylon blend with a modified collar and partial mesh lining, this baseball jacket is the slickest iteration of the style yet
        \vspace{0.9cm}
    \end{subfigure}
    \hspace{0.5cm}
    \begin{subfigure}[b]{0.2\linewidth}
        \centering
        \includegraphics[width=\textwidth]{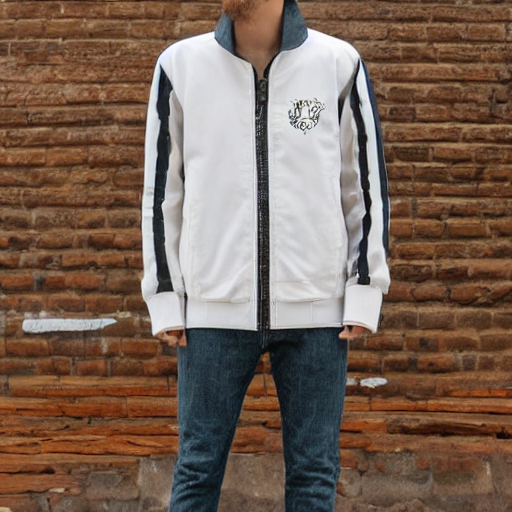}
    \end{subfigure}
    \begin{subfigure}[b]{0.2\linewidth}
        \centering
        \includegraphics[width=\textwidth]{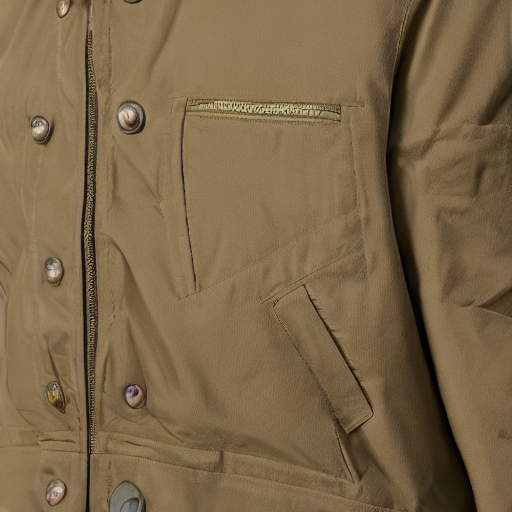}
    \end{subfigure}
    \begin{subfigure}[b]{0.4\linewidth}
        \scriptsize
        \textbf{end\_leaf}: Pants 
    \textbf{gender\_category}: Men
    \textbf{color}: Black-grey
    \textbf{first sentence of description}: This jogger's easy, slouchy silhouette gets a little grit courtesy of its eye-popping print of photorealistic roses
    \vspace{1cm}
    \end{subfigure}
    \hspace{0.5cm}
    \begin{subfigure}[b]{0.2\linewidth}
        \centering
        \includegraphics[width=\textwidth]{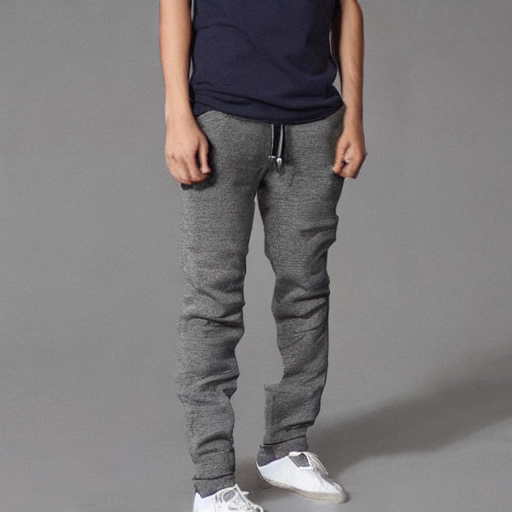}
    \end{subfigure}
    \begin{subfigure}[b]{0.2\linewidth}
        \centering
        \includegraphics[width=\textwidth]{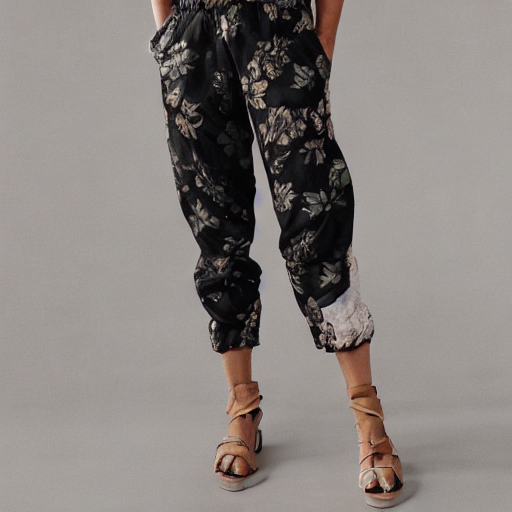}
    \end{subfigure}
    \begin{subfigure}[b]{0.4\linewidth}
        \scriptsize
        \textbf{end\_leaf}: Shirts Polos
    \textbf{gender\_category}: Men
    \textbf{color}: Coral
    \textbf{first sentence of description}: Constructed from cotton for a classic fit, this lightweight shirt features buttoned chest pockets
    \vspace{1cm}
    \end{subfigure}
    \hspace{0.5cm}
    \begin{subfigure}[b]{0.2\linewidth}
        \centering
        \includegraphics[width=\textwidth]{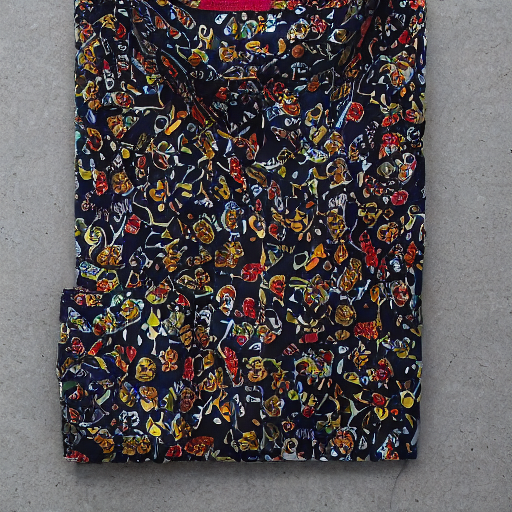}
    \end{subfigure}
    \begin{subfigure}[b]{0.2\linewidth}
        \centering
        \includegraphics[width=\textwidth]{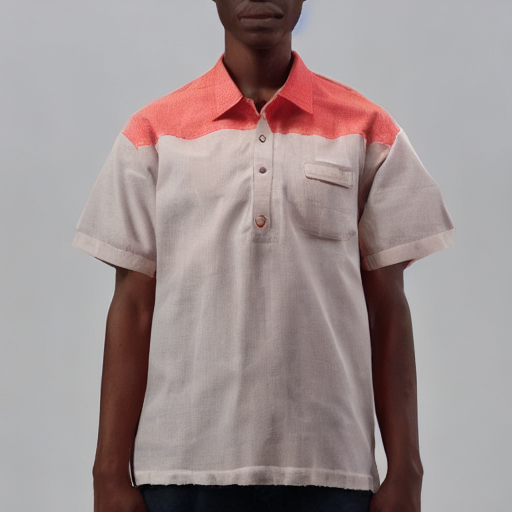}
    \end{subfigure}
    \begin{subfigure}[b]{0.4\linewidth}
        \scriptsize
        \textbf{end\_leaf}: Shorts
    \textbf{gender\_category}: Men
    \textbf{color}: Grey
    \textbf{first sentence of description}: Crafted from speckled French terry, this sharper-than -average pair of sweatshorts is outfitted with a mock fly and three shiny zip pockets (two in front, one in back), ideal for lounging around or winning triathalons (just kidding)
    \vspace{0.7cm}
    \end{subfigure}
    \hspace{0.5cm}
    \begin{subfigure}[b]{0.2\linewidth}
        \centering
        \includegraphics[width=\textwidth]{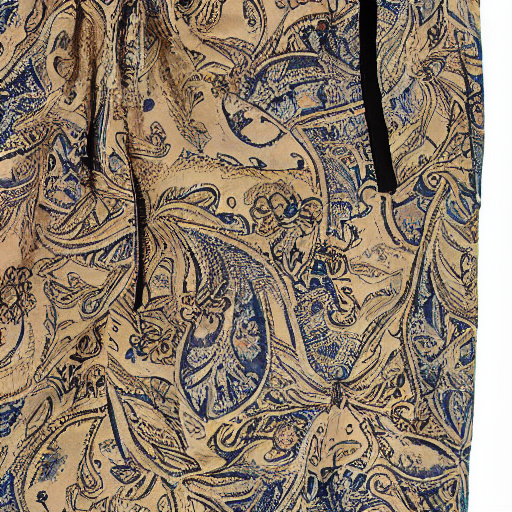}
    \end{subfigure}
    \begin{subfigure}[b]{0.2\linewidth}
        \centering
        \includegraphics[width=\textwidth]{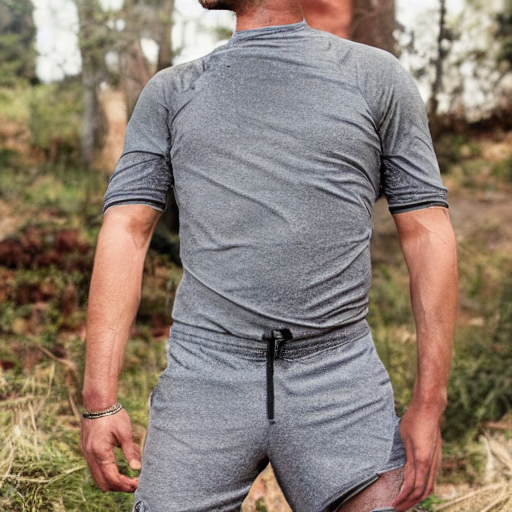}
    \end{subfigure}
    \begin{subfigure}[b]{0.4\linewidth}
        \scriptsize
        \textbf{end\_leaf}:  Blouses Shirts
    \textbf{gender\_category}: Women
    \textbf{color}: Rust
    \textbf{first sentence of description}: Effortlessly ethereal and romantic, this cutout-shoulder top is what dream closets are made of
    \vspace{1cm}
    \caption{Metadata}
    \end{subfigure}
    \hspace{0.5cm}
    \begin{subfigure}[b]{0.2\linewidth}
        \centering
        \includegraphics[width=\textwidth]{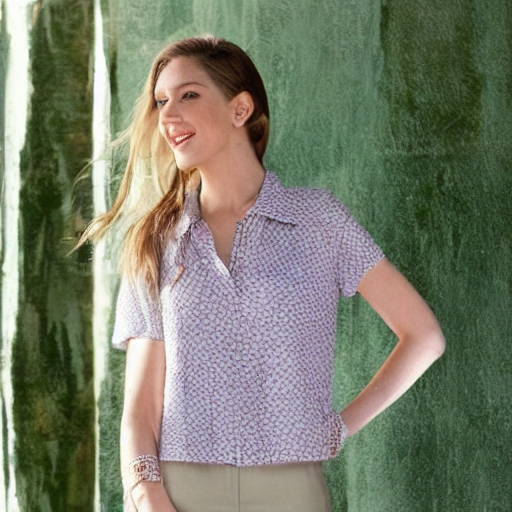}
        \caption{Prompt 3}
    \end{subfigure}
    \begin{subfigure}[b]{0.2\linewidth}
        \centering
        \includegraphics[width=\textwidth]{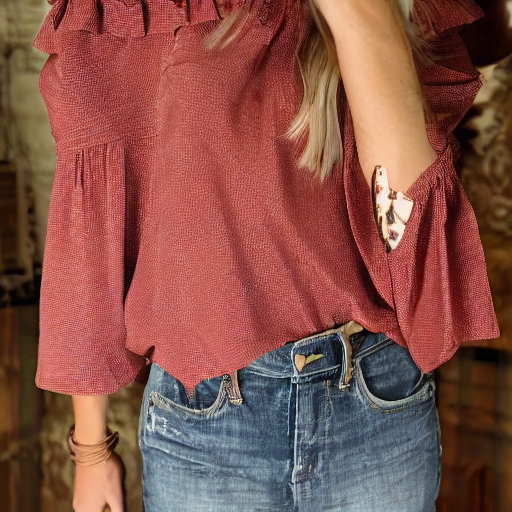}
        \caption{Prompt 4}
    \end{subfigure}
    \end{center}
    \caption{Generated images with Vanilla SD model to determine prompt.}
    \label{fig:t2i_sup_prompt_eng_imgs}
\end{figure*}

\renewcommand{\thetable}{Supp-3}
\begin{table}[!t]
  	\begin{center}
    \caption{Prompts evaluated for text-to-image generation experiment. Prompt structures varied slightly due to available metadata across datasets.}
    \begin{small}
    \begin{tabular}{c!{\vrule width 2pt}c|c}
      	\toprule
      	\textbf{Prompt ID} & \textbf{Dataset} & \textbf{Prompt Structure} \\
      	\midrule
      	1 & LGS & \{brand\} \{title\} in the style of e-commerce \\
        & DeepFashion & \{first sentence of description\} in the style of e-commerce  \\ 
      	\hline 
      	2 & LGS   & \{end\_leaf\} advertisement for a \{title\} from \{brand\} \\ %
        & DeepFashion & \{end\_leaf\} advertisement for a \{first sentence of description\}\\
       	\hline 
      	3 & LGS & \{brand\} \{end\_leaf\} \{title\} \{description\} \\
        & DeepFashion & \{end\_leaf\} \{description\} \{gender\_category\} \{color\}  \\ %
      	\hline 
    	4 & LGS & a photo of \{brand\} \{end\_leaf\} \{title\}, e-commerce \\
        & DeepFashion & a photo of \{end\_leaf\} \{color\} \{first sentence of description\}, e-commerce  \\
      	\bottomrule
  	\end{tabular}
    \end{small}
    \label{tab:sd_prompt_examples}
  	\end{center}
\end{table}

\renewcommand{\thefigure}{Supp-8}
\begin{figure*}[!tb]
    \begin{center}
    \begin{subfigure}[b]{0.22\linewidth}
        \scriptsize{a photo of Blouses Shirts Tomato Love 21 - A woven cami featuring a pleated front and crossback strap detail in the back, e-commerce} 
        \vspace{12mm}
    \end{subfigure}
    \begin{subfigure}[b]{0.22\linewidth}
        \centering
        \includegraphics[width=\textwidth]{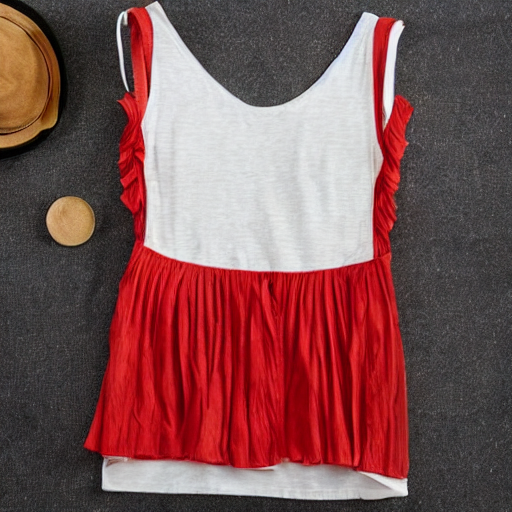}
    \end{subfigure}
    \begin{subfigure}[b]{0.22\linewidth}
        \centering
        \includegraphics[width=\textwidth]{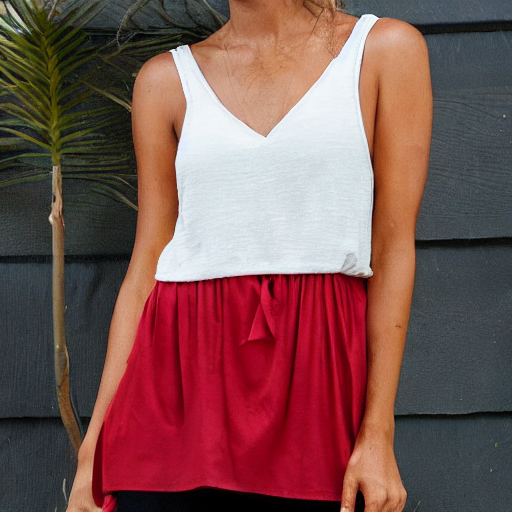}
    \end{subfigure} \\
    \begin{subfigure}[b]{0.22\linewidth}
        \scriptsize{a photo of Jackets Vests Khaki Made in a cotton-nylon blend with a modified collar and partial mesh lining, this baseball jacket is the slickest iteration of the style yet, e-commerce}
        \vspace{9mm}
    \end{subfigure}
    \begin{subfigure}[b]{0.22\linewidth}
        \centering
        \includegraphics[width=\textwidth]{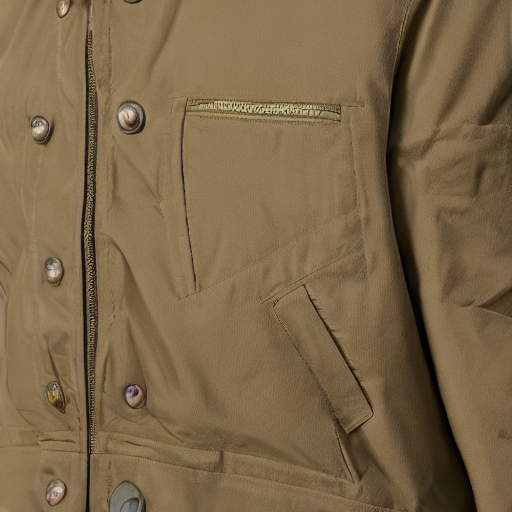}
    \end{subfigure}
    \begin{subfigure}[b]{0.22\linewidth}
        \centering
        \includegraphics[width=\textwidth]{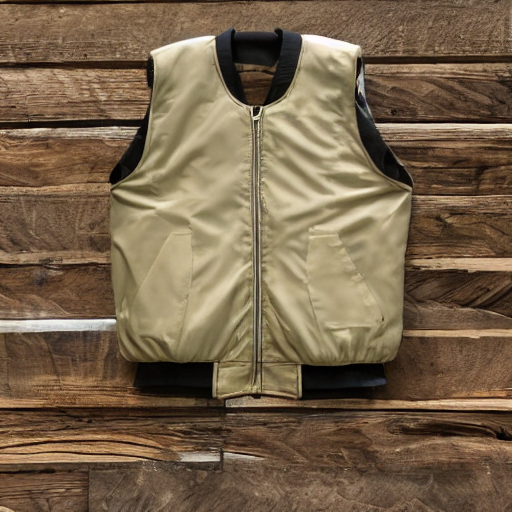}
    \end{subfigure} \\
    \begin{subfigure}[b]{0.22\linewidth}
        \scriptsize{a photo of Shirts Polos Coral Constructed from cotton for a classic fit, this lightweight shirt features buttoned chest pockets, e-commerce}
        \vspace{12mm}
    \end{subfigure}
    \begin{subfigure}[b]{0.22\linewidth}
        \centering
        \includegraphics[width=\textwidth]{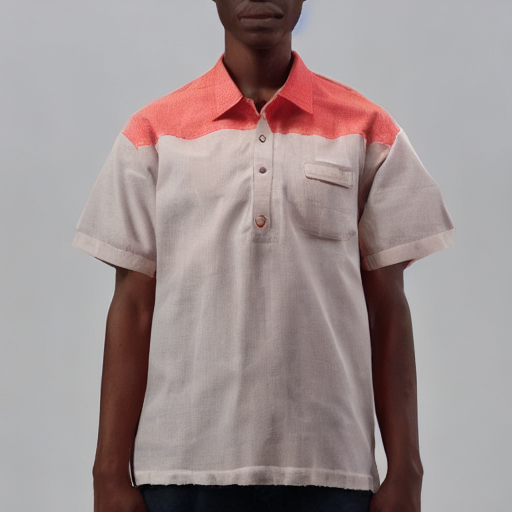}
    \end{subfigure}
    \begin{subfigure}[b]{0.22\linewidth}
        \centering
        \includegraphics[width=\textwidth]{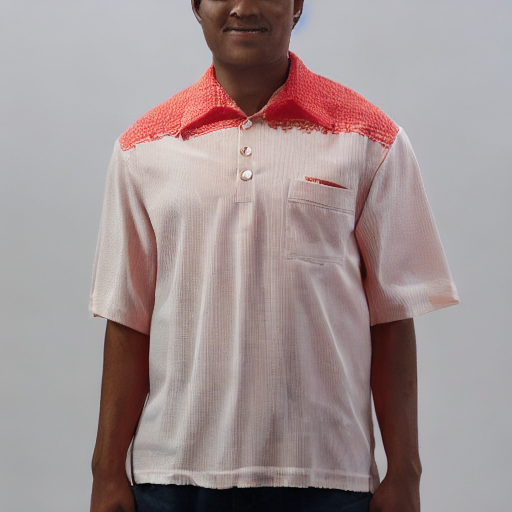}
    \end{subfigure} \\
    \begin{subfigure}[b]{0.22\linewidth}
        \scriptsize{a photo of Shorts Grey Crafted from speckled French terry, this sharper-than-average pair of sweatshorts is outfitted with a mock fly and three shiny zip pockets (two in front, one in back), ideal for lounging around or winning triathalons (just kidding), e-commerce}
        \vspace{6mm}
        \caption{\footnotesize Input Prompt}
    \end{subfigure}
    \begin{subfigure}[b]{0.22\linewidth}
        \centering
        \includegraphics[width=\textwidth]{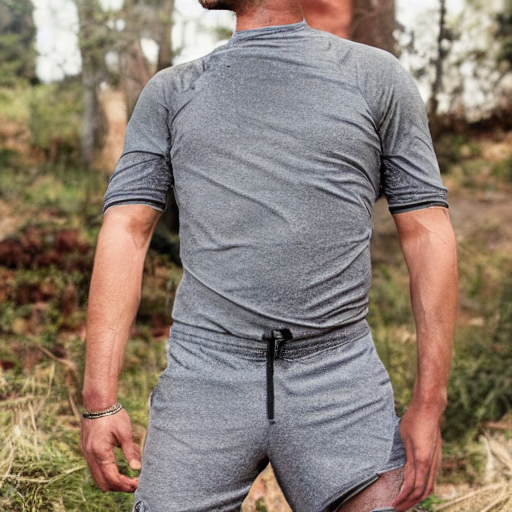}
        \caption{\footnotesize Vanilla SD}
    \end{subfigure}
    \begin{subfigure}[b]{0.22\linewidth}
        \centering
        \includegraphics[width=\textwidth]{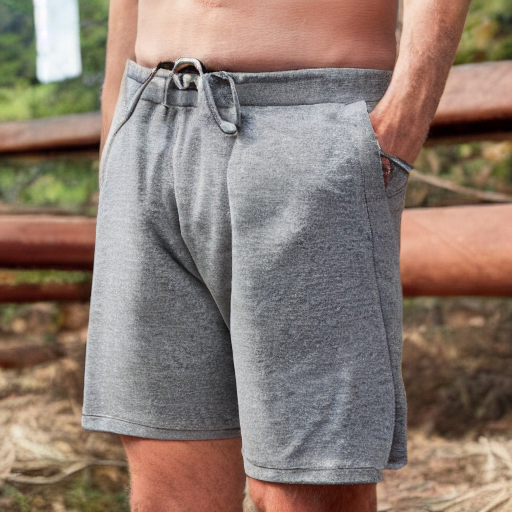}
        \caption{\footnotesize LGS-117}
    \end{subfigure}
    \caption{Additional qualitative examples of the Vanilla SD vs LGS-117 fine-tuned SD model on DeepFashion InShop dataset.}
    \vspace{-0.5cm}
    \label{fig:t2i_sup_inshop_imgs}
    \end{center}
\end{figure*}

\renewcommand{\thefigure}{Supp-9}
\begin{figure*}[!tb]
    \begin{center}
    \begin{subfigure}[b]{0.22\linewidth}
        \scriptsize{a photo of ana silver co. earrings rainbow moonstone earrings 3/4" (925 sterling silver) earr415021}
        \vspace{12mm}
    \end{subfigure}
    \begin{subfigure}[b]{0.2\linewidth}
        \centering
        \includegraphics[width=\textwidth]{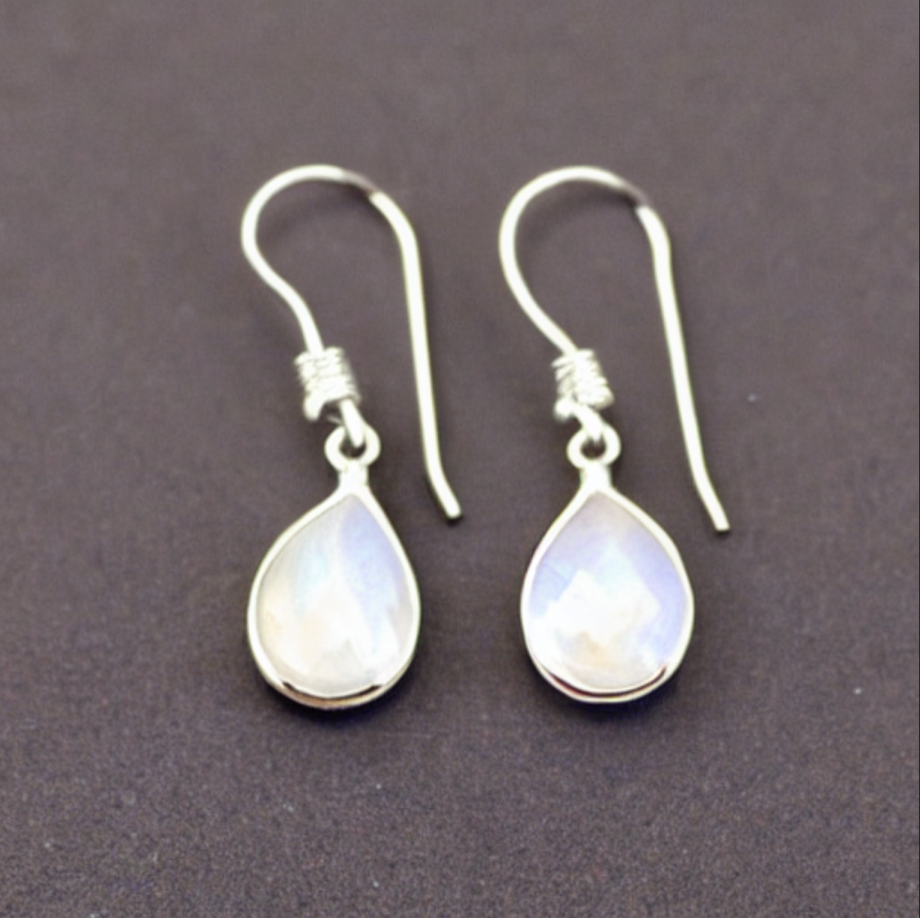}
    \end{subfigure}
    \begin{subfigure}[b]{0.2\linewidth}
        \centering
        \includegraphics[width=\textwidth]{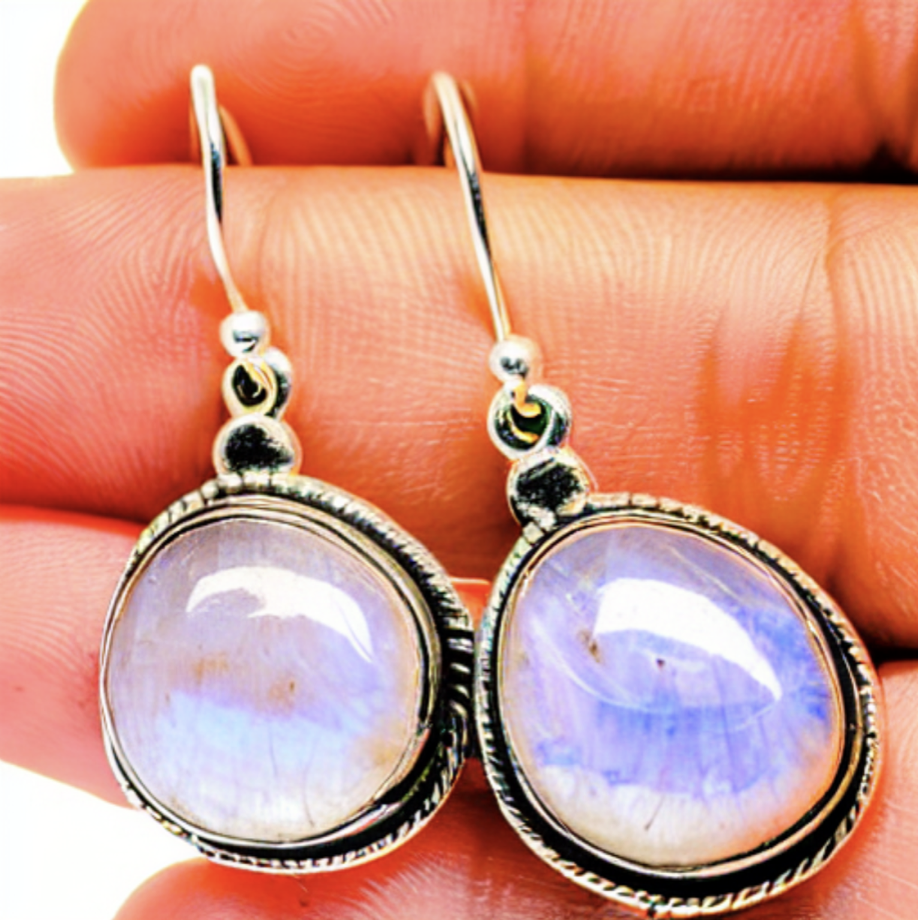}
    \end{subfigure}
    
    \begin{subfigure}[b]{0.22\linewidth}
        \scriptsize{a photo of vans vault vans vault old skool lx - croc skin/flame}
        \vspace{15mm}
    \end{subfigure}
    \begin{subfigure}[b]{0.2\linewidth}
        \centering
        \includegraphics[width=\textwidth]{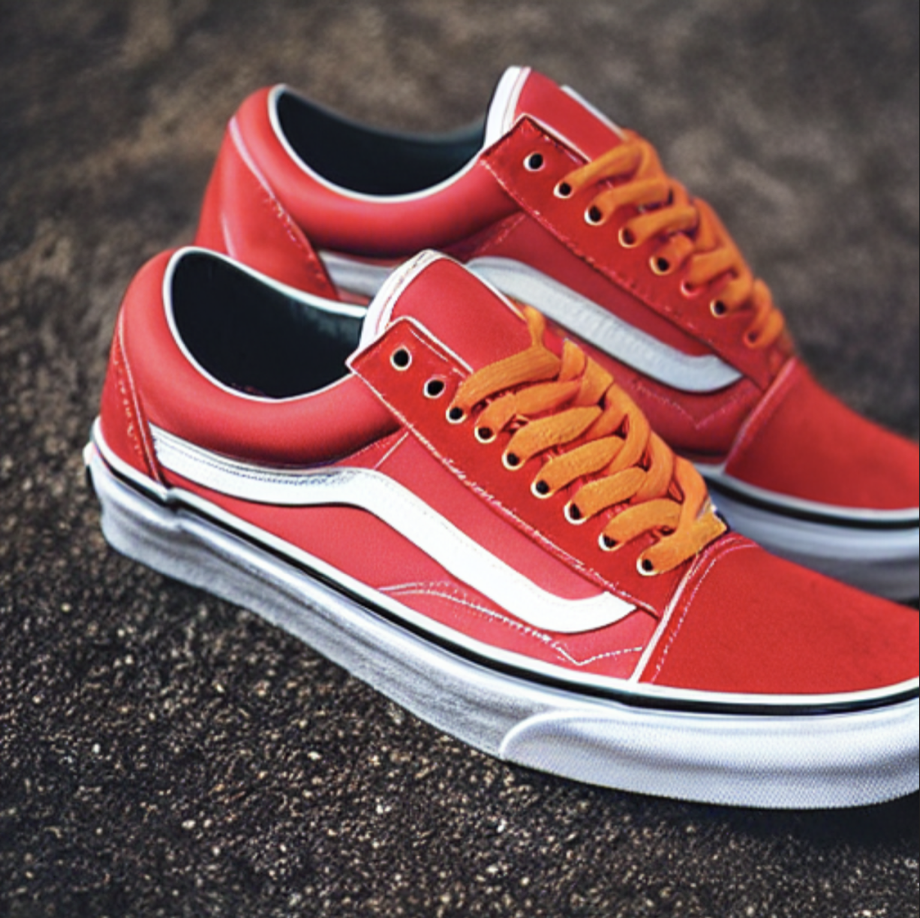}
    \end{subfigure}
    \begin{subfigure}[b]{0.2\linewidth}
        \centering
        \includegraphics[width=\textwidth]{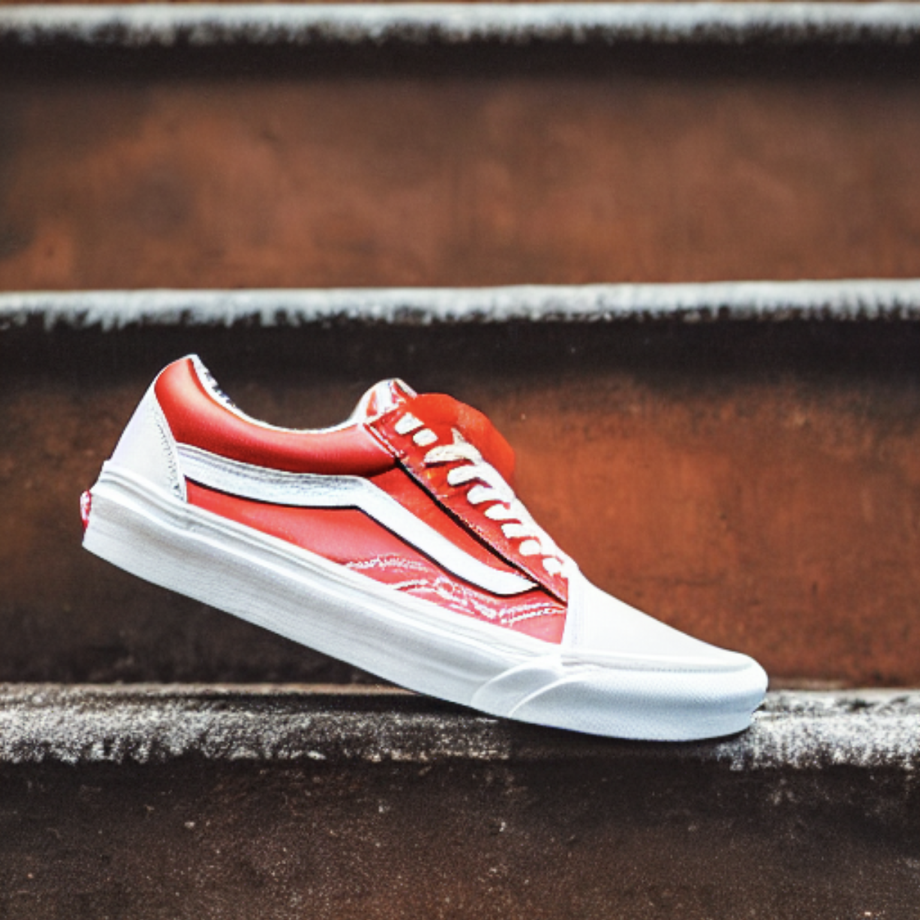}
    \end{subfigure}
    
    \begin{subfigure}[b]{0.22\linewidth}
        \scriptsize{a photo of myconquering conquering unisex black joggers}
        \vspace{15mm}
    \end{subfigure}
    \begin{subfigure}[b]{0.2\linewidth}
        \centering
        \includegraphics[width=\textwidth]{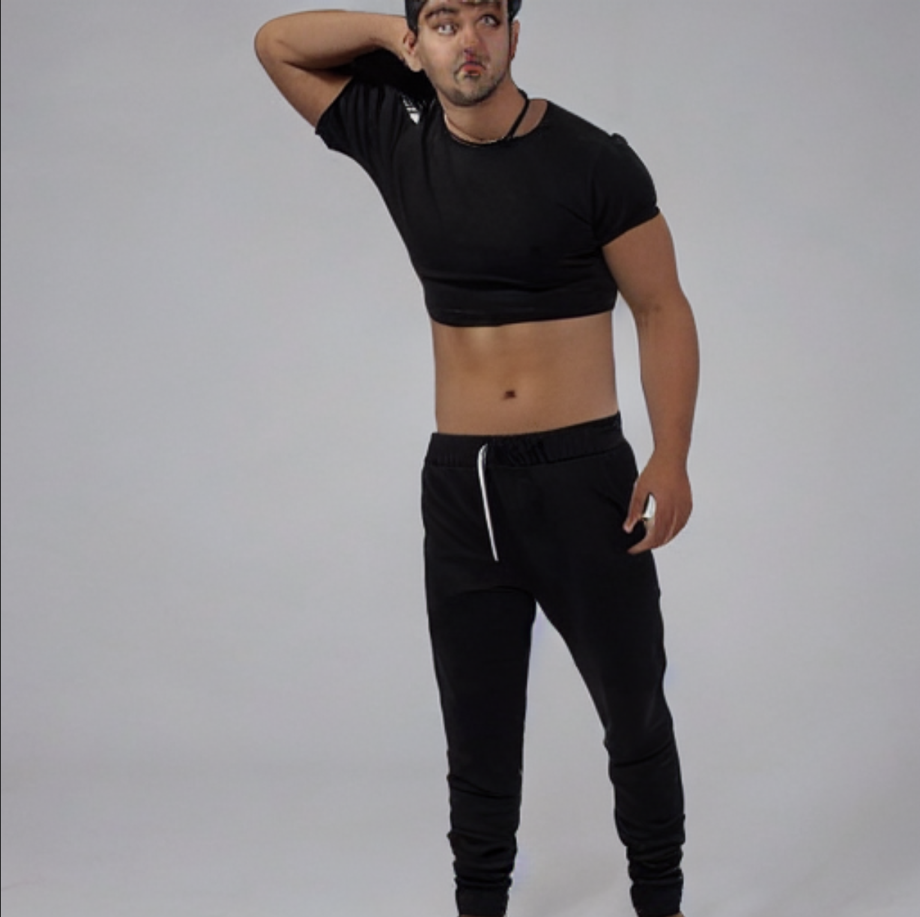}
    \end{subfigure}
    \begin{subfigure}[b]{0.2\linewidth}
        \centering
        \includegraphics[width=\textwidth]{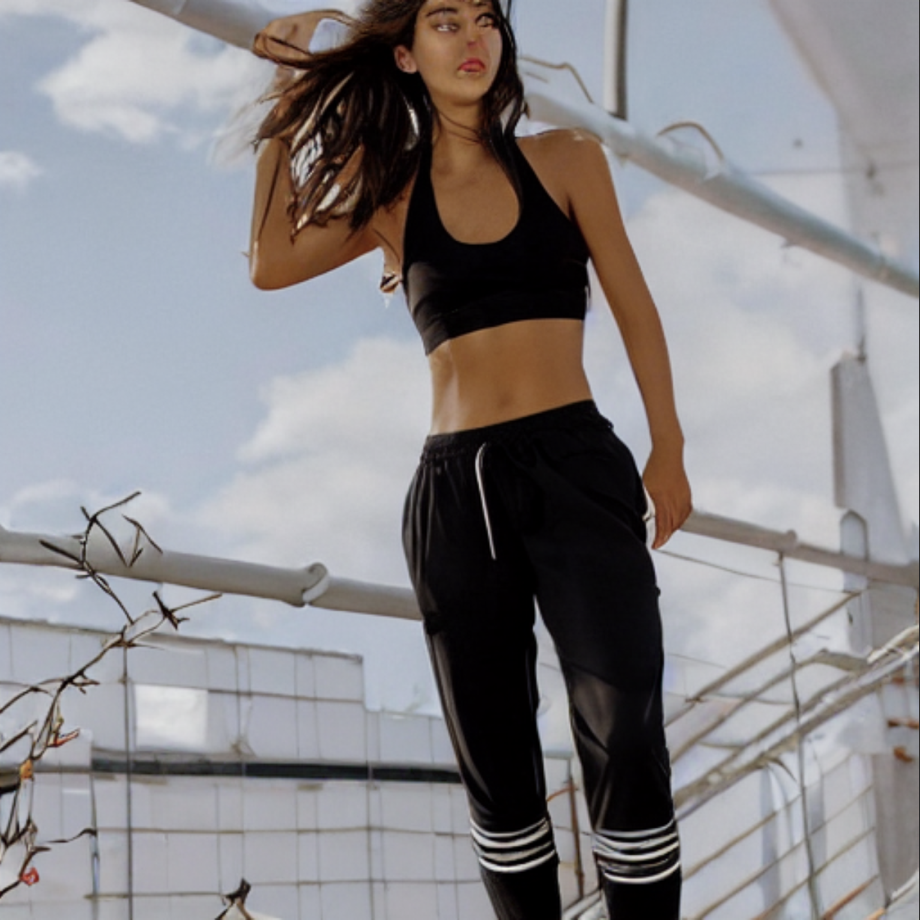}
    \end{subfigure}
    
    \begin{subfigure}[b]{0.22\linewidth}
        \scriptsize{a photo of chopard cat eye unisex sunglasses}
        \vspace{15mm}
    \end{subfigure}
    \begin{subfigure}[b]{0.2\linewidth}
        \centering
        \includegraphics[width=\textwidth]{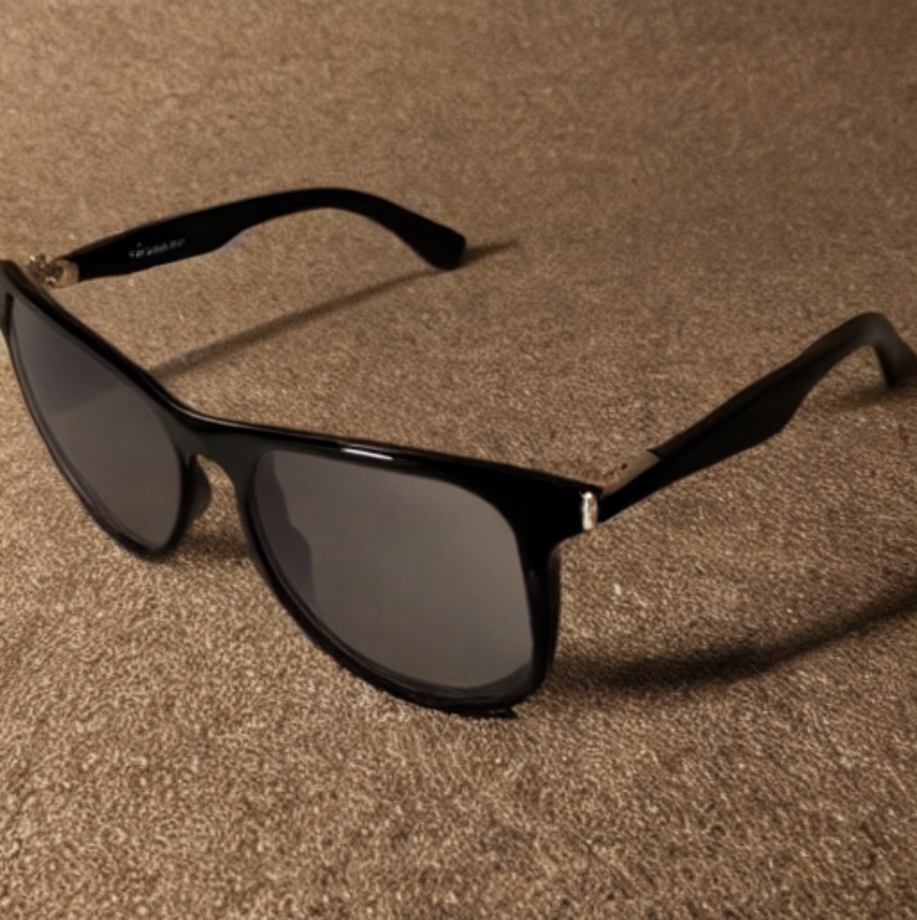}
    \end{subfigure}
    \begin{subfigure}[b]{0.2\linewidth}
        \centering
        \includegraphics[width=\textwidth]{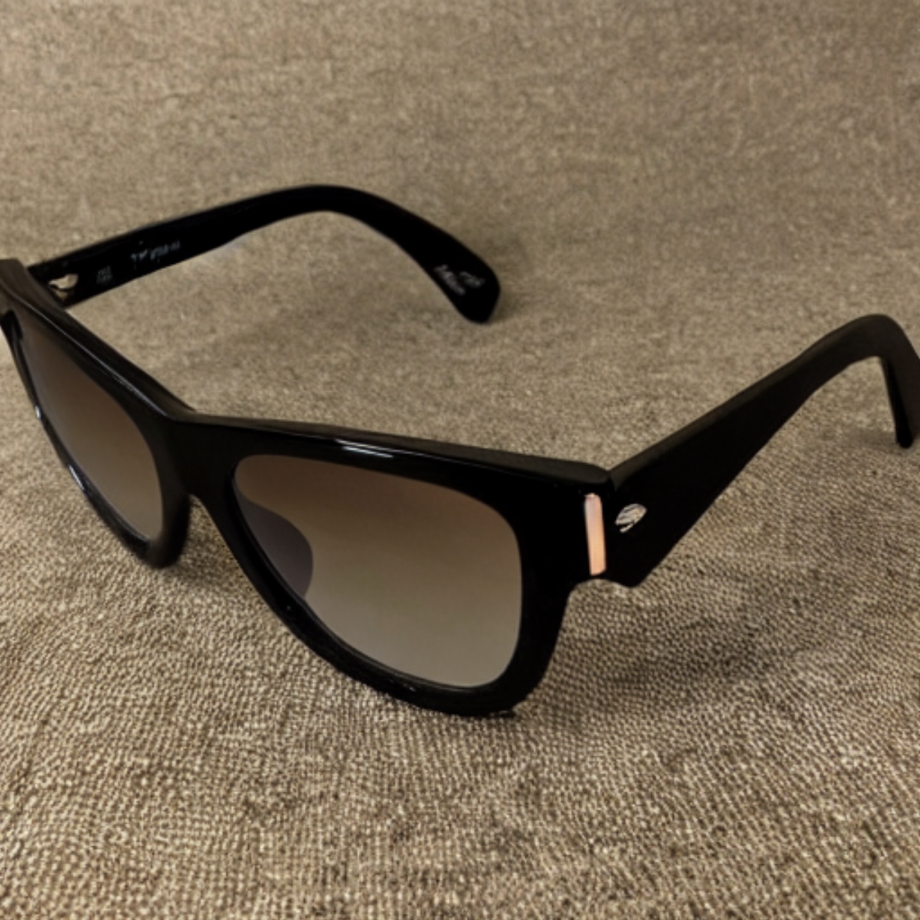}
    \end{subfigure}
    
    \begin{subfigure}[b]{0.22\linewidth}
        \scriptsize{a photo of invicta bracelets elements men's bracelet}
        \vspace{15mm}
    \end{subfigure}
    \begin{subfigure}[b]{0.2\linewidth}
        \centering
        \includegraphics[width=\textwidth]{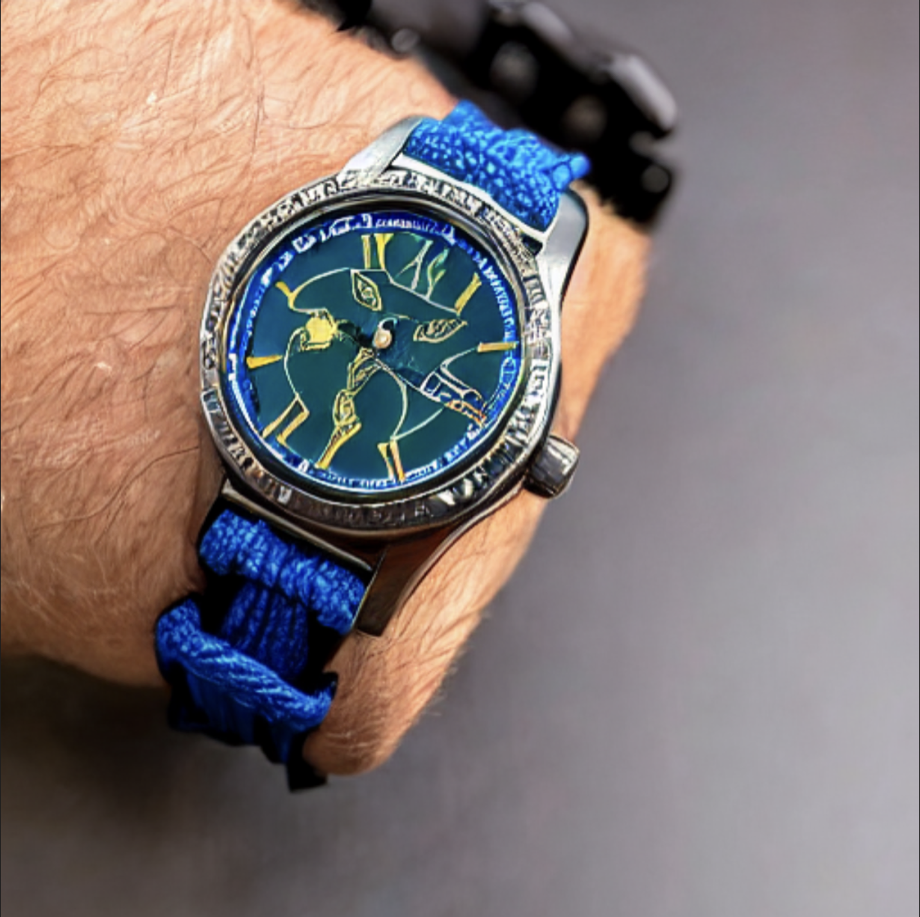}
    \end{subfigure}
    \begin{subfigure}[b]{0.2\linewidth}
        \centering
        \includegraphics[width=\textwidth]{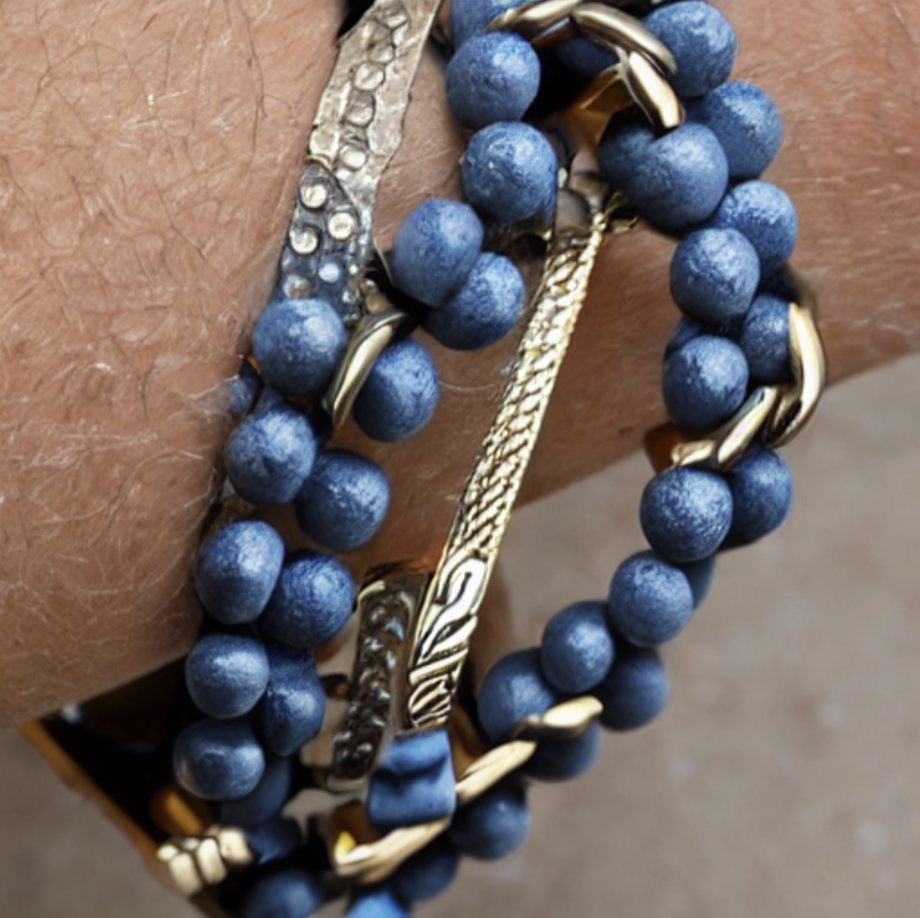}
    \end{subfigure}
    
    \begin{subfigure}[b]{0.22\linewidth}
        \scriptsize{a photo of wristwatchstraps.co smart watch accessories bumper cover+glass for apple watch - lilac 21 - 38mm}
        \vspace{1cm}
        \caption{\scriptsize Input Prompt}
    \end{subfigure}
    \begin{subfigure}[b]{0.2\linewidth}
        \centering
        \includegraphics[width=\textwidth]{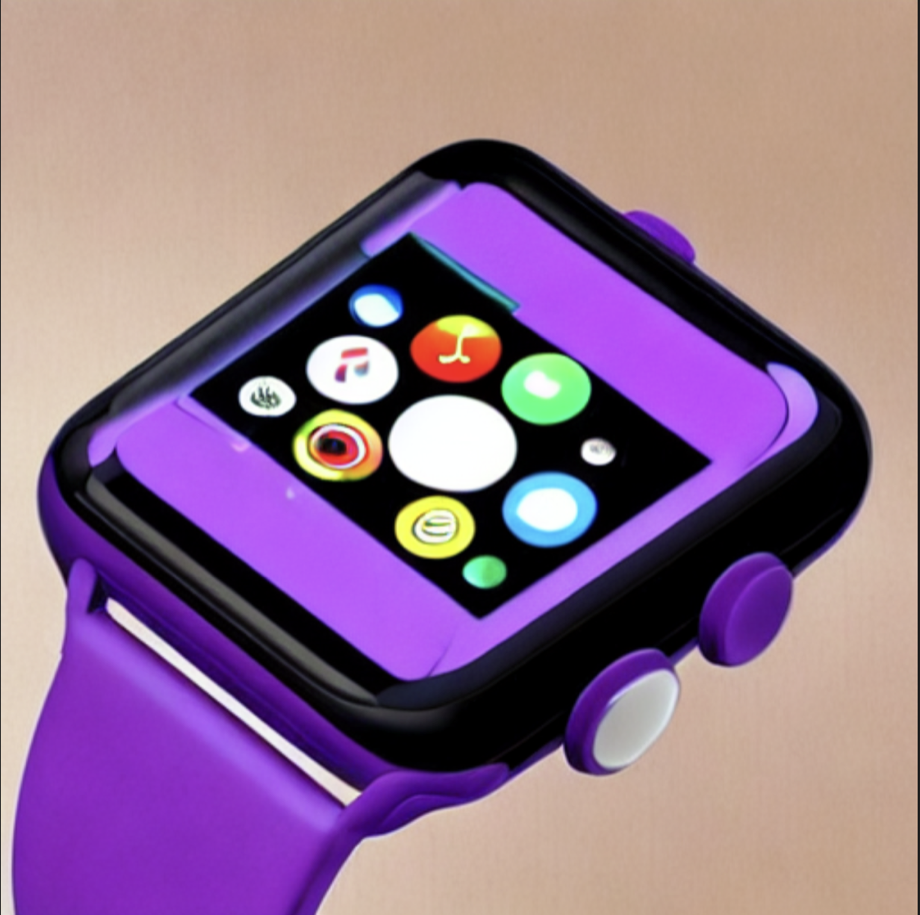}
        \caption{\scriptsize Vanilla SD}
    \end{subfigure}
    \begin{subfigure}[b]{0.2\linewidth}
        \centering
        \includegraphics[width=\textwidth]{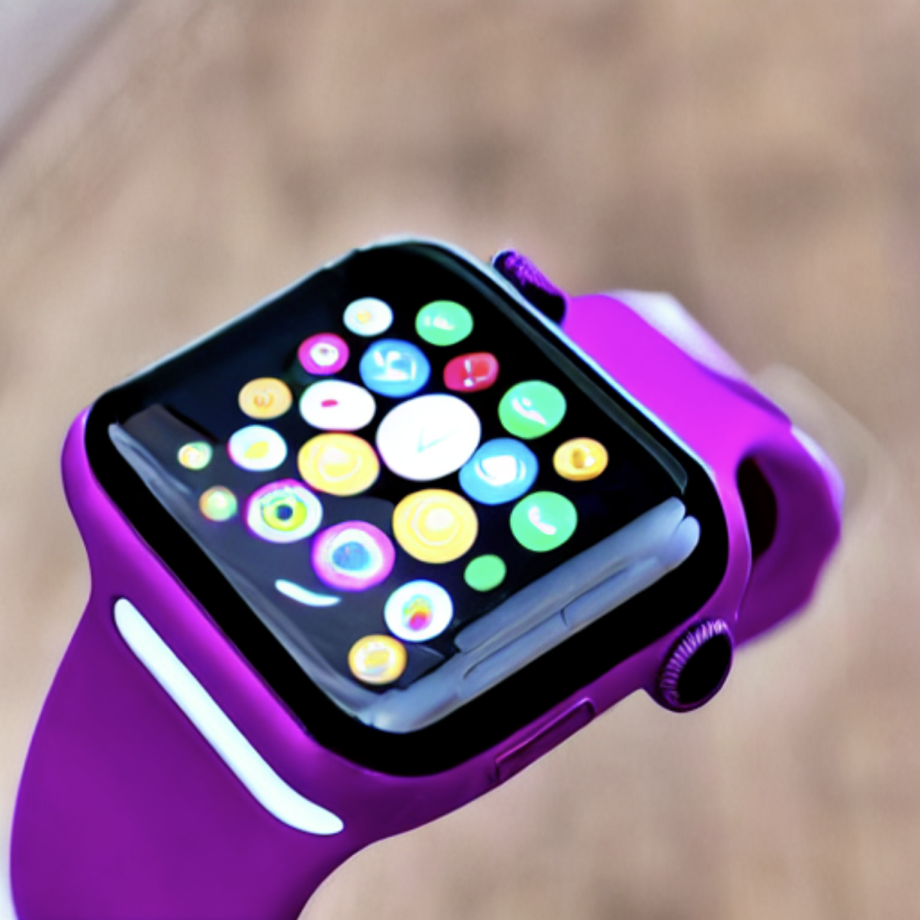}
        \caption{\scriptsize LGS-117}
    \end{subfigure}
    
    \caption{Additional qualitative examples of the Vanilla SD vs LGS-117 fine-tuned SD model on the LGS dataset.}
    \vspace{-0.5cm}
    \label{fig:t2i_sup}
    \end{center}
\end{figure*}

\end{document}